\documentclass[letterpaper]{article} 
\usepackage[draft]{AuthorKit26/AnonymousSubmission/LaTeX/aaai2026} 
\usepackage{times}  
\usepackage{helvet}  
\usepackage{courier}  
\usepackage[hyphens]{url}  
\usepackage{graphicx} 
\usepackage{natbib}  
\usepackage{caption} 
\usepackage{algorithm}
\usepackage{algorithmic}
\usepackage{tabularx}
\usepackage{multirow}
\usepackage{enumitem}
\newcolumntype{Y}{>{\centering\arraybackslash}X}
\newcolumntype{N}{>{\hsize=.975\hsize\centering\arraybackslash}X}  
\newcolumntype{B}{>{\hsize=1.25\hsize\centering\arraybackslash}X}  
\usepackage{newfloat}
\usepackage{listings}
\DeclareCaptionStyle{ruled}{labelfont=normalfont,labelsep=colon,strut=off} 
\floatstyle{ruled}
\newfloat{listing}{tb}{lst}{}
\floatname{listing}{Listing}
\DeclareCaptionFont{ninept}{\fontsize{10pt}{11pt}\selectfont}
\makeatletter
\def\acro@all{}  
\newcommand{\DeclareAcronym}[3]{%
  \expandafter\gdef\csname acro@short@#1\endcsname{#2}%
  \expandafter\gdef\csname acro@long@#1\endcsname{#3}%
  \global\expandafter\let\csname acro@used@#1\endcsname\relax
  \ifx\acro@all\@empty\gdef\acro@all{#1}\else\g@addto@macro\acro@all{,#1}\fi}
\newcommand{\resetacronyms}{%
  \@for\acro@key:=\acro@all\do{%
    \global\expandafter\let\csname acro@used@\acro@key\endcsname\relax}}
\newcommand{\acs}[1]{\csname acro@short@#1\endcsname}
\newcommand{\acsp}[1]{\csname acro@short@#1\endcsname s}
\newcommand{\ac}[1]{%
  \expandafter\ifx\csname acro@used@#1\endcsname\relax
    \csname acro@long@#1\endcsname\ (\csname acro@short@#1\endcsname)%
    \expandafter\gdef\csname acro@used@#1\endcsname{}%
  \else\csname acro@short@#1\endcsname\fi}
\newcommand{\acp}[1]{%
  \expandafter\ifx\csname acro@used@#1\endcsname\relax
    \csname acro@long@#1\endcsname s\ (\csname acro@short@#1\endcsname s)%
    \expandafter\gdef\csname acro@used@#1\endcsname{}%
  \else\csname acro@short@#1\endcsname s\fi}
\makeatother
\DeclareAcronym{TSLM}{TSLM}{Time Series Language Model}
\DeclareAcronym{RLM}{RLM}{Recursive Language Model}
\DeclareAcronym{LLM}{LLM}{Large Language Model}
\DeclareAcronym{REPL}{REPL}{Read--Eval--Print Loop}
\DeclareAcronym{VLM}{VLM}{Vision Language Model}
\DeclareAcronym{TSFM}{TSFM}{Time Series Foundation Model}
\DeclareAcronym{SFT}{SFT}{Supervised Fine-Tuning}
\DeclareAcronym{GRPO}{GRPO}{Group Relative Policy Optimization}
\DeclareAcronym{OOM}{OOM}{Out of Memory}
\DeclareAcronym{ECG}{ECG}{Electrocardiogram}
\DeclareAcronym{IOU}{IoU}{Intersection over Union}
\DeclareAcronym{pp}{pp}{percentage points}
\DeclareAcronym{RL}{RL}{Reinforcement Learning}
\DeclareAcronym{MCQ}{MCQ}{Multiple Choice Question}

\usepackage[table]{xcolor}  

\usepackage{booktabs}
\usepackage{array}     
\definecolor{ourrow}{gray}{0.90}  
\usepackage{amsmath}   
\usepackage{amssymb}   
\usepackage{tcolorbox}  
\tcbuselibrary{breakable}

\newtcolorbox{promptbox}[1][]{%
  colback=black!3, colframe=black!55, boxrule=0.5pt, arc=2.5mm,
  left=5pt, right=5pt, top=4pt, bottom=4pt,
  fonttitle=\bfseries\small, coltitle=black, colbacktitle=black!10,
  fontupper=\small\ttfamily, breakable, #1}

\title{TimeRLM: Recursive Language Models Enable Precise Anomaly Localization in Long-Context Time-Series}
\author{
    \large
    \newcommand{\authorsep}{\hspace{0.7em}}%
    \mbox{Nicolas Zumarraga\textsuperscript{\rm 1}}\thanks{Corresponding author -- nzumarraga@ethz.ch}\authorsep
    \mbox{Lorenzo Steno\textsuperscript{\rm 1,3}}\authorsep
    \mbox{Ning Wang\textsuperscript{\rm 1}}\authorsep
    \mbox{Max Rosenblattl\textsuperscript{\rm 2}}\authorsep
    \mbox{Thomas Kaar\textsuperscript{\rm 2}}\authorsep
    \mbox{Maxwell A.\ Xu\textsuperscript{\rm 4,5}}\authorsep
    \mbox{Kevin O'Sullivan\textsuperscript{\rm 1}}\authorsep
    \mbox{Markus Kreft\textsuperscript{\rm 1}}\authorsep
    \mbox{Elgar Fleisch\textsuperscript{\rm 1,6,7}}\authorsep
    \mbox{Paul Schmiedmayer\textsuperscript{\rm 2}}\authorsep
    \mbox{Patrick Langer\textsuperscript{\rm 1,2}}\thanks{Equal contribution as senior authors.}\authorsep
    \mbox{Robert Jakob\textsuperscript{\rm 1}\footnotemark[2]}
}
\affiliations{
    \small
    \newcommand{\affilsep}{\hspace{0.9em}}%
    \mbox{\textsuperscript{\rm 1}Agentic Systems Lab, ETH Z\"urich}\affilsep
    \mbox{\textsuperscript{\rm 2}Stanford University}\affilsep
    \mbox{\textsuperscript{\rm 3}Seldon Technologies}\affilsep
    \mbox{\textsuperscript{\rm 4}University of Illinois Urbana-Champaign}\affilsep
    \mbox{\textsuperscript{\rm 5}Google}\\
    \mbox{\textsuperscript{\rm 6}Centre for Digital Health Interventions, ETH Z\"urich}\affilsep
    \mbox{\textsuperscript{\rm 7}Centre for Digital Health Interventions, University of St.~Gallen}
}

\begin{document}

\maketitle

\begin{abstract}
Precise anomaly localization over long-context time-series is a crucial task in monitoring applications across clinical care, industrial operations, financial services, and logistics, where brief evidence may hide inside long spans of high-frequency data.
\acp{TSLM} are able to ingest time series data and verbalize findings on anomalies in natural language; however recent benchmarks report a decrease in retrieval performance at long time horizons, mirroring failure modes for long-context retrieval in text, vision, and audio.
In the text domain, \acp{RLM} can recover much of this lost performance by keeping context external to the \ac{LLM} rather than placing it in the input prompt, allowing the model to query and manipulate it through code.
%
We present \textbf{TimeRLM}, an \ac{RLM} formulation for time-series that sequentially manipulates the signal using code and vision capabilities.
To assess performance, we further introduce \textbf{\mbox{AnomalyXL}}, a synthetic long-context anomaly localization benchmark with programmatically injected anomalies that require precise retrieval. We implement five different task categories and two variants; AnomalyXL-MCQ and AnomalyXL-Localize. 
TimeRLM outperforms every evaluated \ac{TSLM} and frontier single-pass model on four of the five AnomalyXL-Localize tasks, reaching $0.682$ \ac{IOU} on localization and $0.745$ on classify-with-evidence, compared with at most $0.329$ and $0.072$ across all baselines.
To further improve performance, we train TimeRLM using reinforcement learning and re-evaluate it on the same datasets. The resulting model further improves performance and outperforms \acp{TSLM} on precise retrieval tasks while using an \ac{LLM} backbone up to eight times smaller. It also requires approximately one-third as many agent interaction turns as the untrained base model to produce a final answer. When evaluated on previously unseen real-world \ac{ECG}, sleep and software observability recordings, the post-trained TimeRLM retains or improves performance, surpassing \acp{TSLM} despite being trained exclusively on synthetic data.
%
Our findings suggest recursive interaction with time-series as an effective approach for long-horizon temporal reasoning and retrieval. \looseness -1
\end{abstract}
\resetacronyms

\begin{links}
    \newcommand{\linklogo}[1]{%
      \raisebox{-0.28\height}{\includegraphics[height=1.15em]{#1}}}
    \par\linklogo{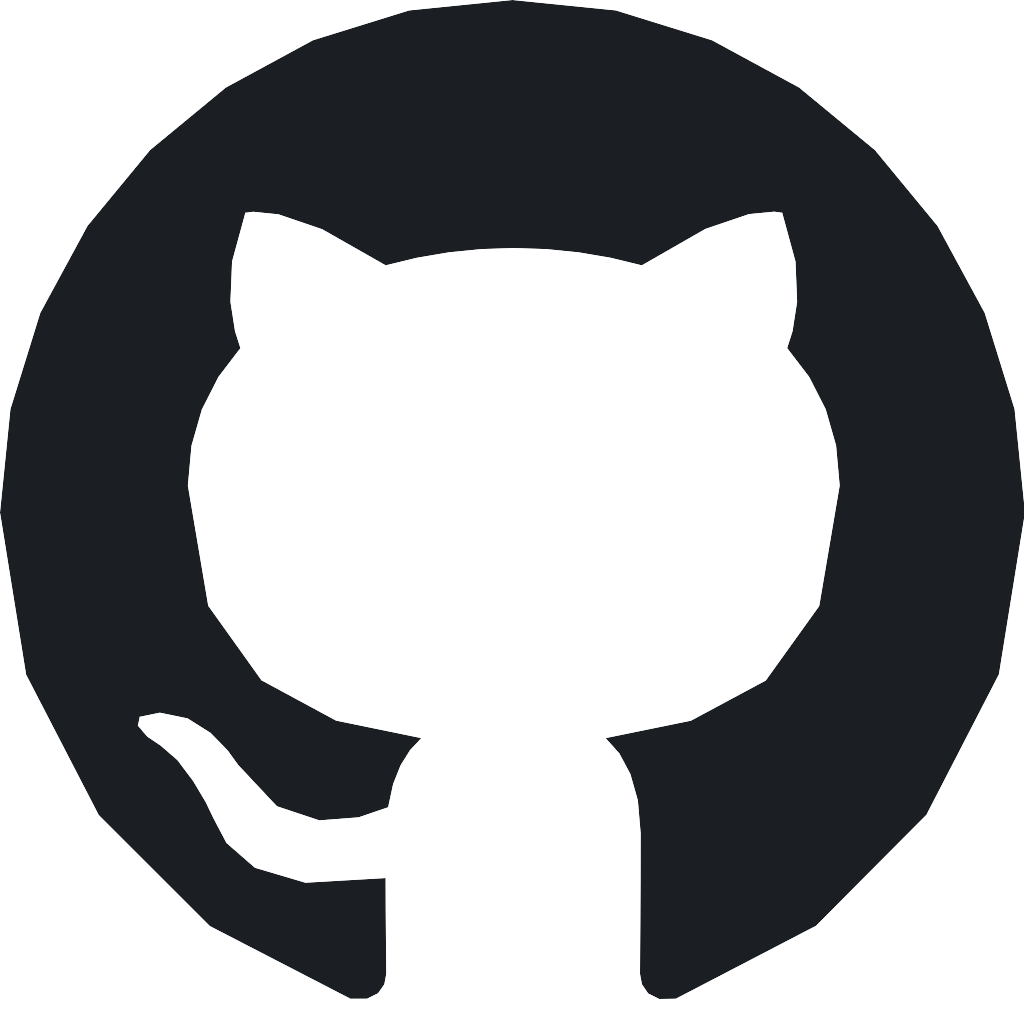}\hspace{0.4em}%
    \linklogo{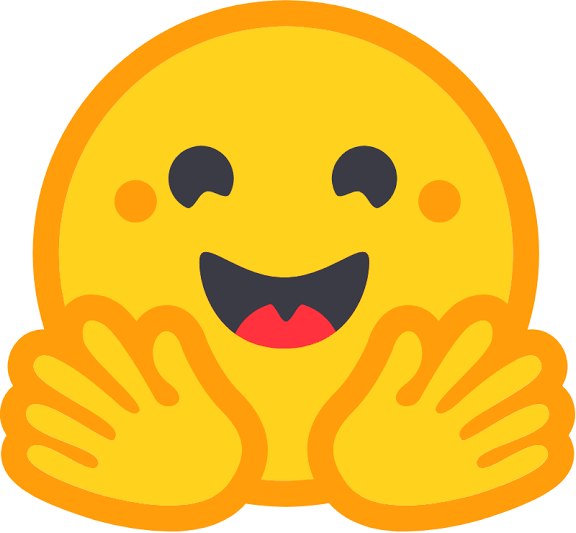}\hspace{0.5em}---\hspace{0.5em}%
    \url{https://github.com/OpenTSLM/TimeRLM}
\end{links}


\section{Introduction}
\label{sec:introduction}

%

In healthcare, manufacturing, software operations, financial services, and logistics, sensors routinely produce long continuous data streams~\cite{yeche2021hirid,yuan2024oxwearables}. These streams may span hours of intensive-care vital signs~\citep{yeche2021hirid}, days of wearable activity~\citep{yuan2024oxwearables,chan2024capture24}, or years of energy telemetry~\citep{kelly2015ukdale}. Yet the events that matter within this data are often brief anomalies—such as a few abnormal beats, a transient arousal, or a short latency spike—buried in otherwise conventional signals.
Identifying these events requires models that can efficiently scan long time-series, localize relevant intervals, and explain why they are unusual~\cite{Su2024LargeLMAnomaly}. Early work applied \acp{LLM} to anomaly detection and reasoning, but \acp{LLM} struggle to interpret time-series data~\cite{langer2026opentslm}. Recent advances in multimodal modeling have enabled \acp{TSLM} to process raw time-series as a native modality, making them a promising approach for detecting anomalies and describing their characteristics in natural language~\cite{zumarraga2026tshaystack, wang2024chattimeunifiedmultimodaltime, langer2026opentslm}.

\begin{figure*}[t]
\centering
\includegraphics[width=0.98\textwidth]{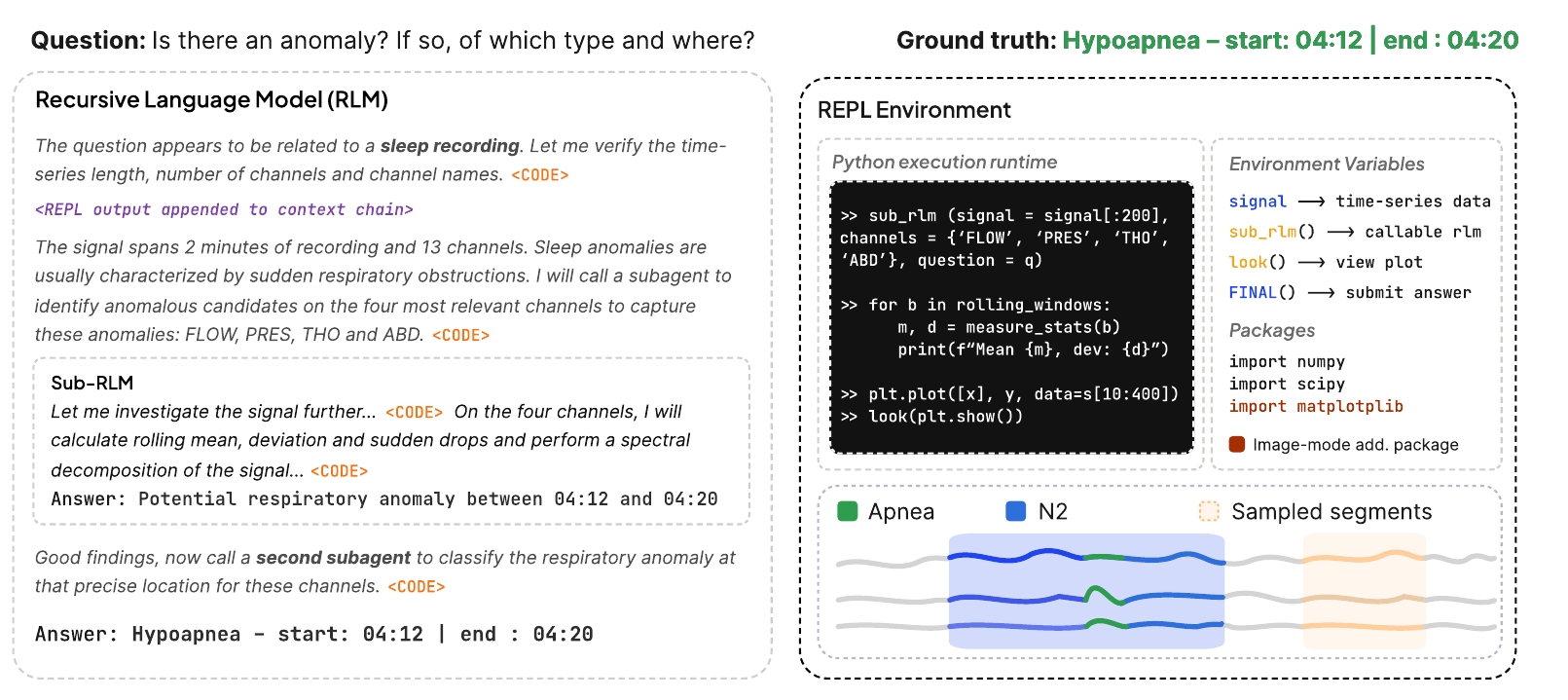}
\caption{\textbf{TimeRLM interacts with time-series through code.}
The orchestrator reasons in natural language and emits Python into a sandboxed \ac{REPL}
(right), where the signal is an environment variable. Sub-RLMs can be invoked.}
\label{fig:timerlm}
\end{figure*}

Different \ac{TSLM} architectures have been proposed to fuse time-series, for example using full-resolution patch
adapters~\citep{wang2024chattimeunifiedmultimodaltime,Xie_2025}, compression of time-series into fixed latent vectors~\citep{langer2026opentslm,wang2025itformer}, and time-series--vision hybrids using rendered
plots~\citep{xie2026arfbench}. While architecturally different, all ingest time-series into context and answer in a single forward pass: the recording is encoded once, at fixed or compressed size, and the model answers from this single view.
In evaluation, time-series question-answering benchmarks have measured model performance on short context anomaly classification, or reduced long horizon localization tasks to multiple-choice questions~\citep{cai2024timeseriesexam,xie2026arfbench}, side-stepping the precise temporal retrieval that long, real-world signals require.
Recent evaluations on long-context \ac{TSLM} retrieval have exposed performance limitations of single-pass \acp{TSLM}, where precise retrieval performance collapses
at increased time-series lengths, leading to TSLMs missing fine-grained events or anomalies~\citep{zumarraga2026tshaystack}. 
This finding aligns with similar long-context retrieval failure modes in text~\citep{kamradt2023needle,liu2023lostmiddle,kuratov2024babilongtestinglimitsllms}, vision~\citep{wang2024needle,song2024milebench}, and audio~\citep{he2025audiomarathon}.

Recent agentic search paradigms employ tool use~\citep{yao2023react, wang2024videoagent} and RL-trained search~\citep{jin2025searchr,chen2026research} to improve performance by letting
the model interact with the input over multiple turns instead of ingesting context in a single pass.
To that end, \acp{RLM} store the full content as a variable rather than placing it in the model’s context window, allowing it to query and manipulate it through code~\citep{zhang2026recursivelanguagemodels}.
Recent work in agentic time-series reasoning has proposed analyzing signals through similar approaches, using tool calls and learned segment selection~\citep{zhao2025timeseriesscientistgeneralpurposeaiagent,messica2026artist},
showcasing the potential of avoiding full context ingestion for time-series reasoning. 
However, a general, training-free
formulation with direct access to the raw signal that works over long-context data is yet to be proposed for time-series reasoning. 
We therefore introduce \textbf{TimeRLM}, an \ac{RLM} formulation that enables manipulating time-series signals as external context, querying them over multiple turns in a sandboxed Python environment, and
reasoning through code and vision to submit a structured answer grounded in the underlying signal. 
To evaluate TimeRLM, we additionally introduce \textbf{AnomalyXL}, an open-source benchmark measuring precise anomaly localization in long horizon time-series spanning two evaluation corpora, \textbf{AnomalyXL-MCQ} and \textbf{AnomalyXL-Localize}. 
AnomalyXL is, to our knowledge, the first synthetic long-context anomaly benchmark released together with code implementations for the generator and \ac{RL} environment. We provide configurable difficulty for generating training and evaluation data and tasks.
In addition to evaluating on AnomalyXL, we further evaluate TimeRLM performance on the anomaly-property tier of ARFBench~\citep{xie2026arfbench}, a multiple-choice benchmark built from real production telemetry. We also assess zero-shot transfer to a generalization set built from real \ac{ECG}~(LTAF dataset)~\citep{petrutiu2007ltaf} and sleep~\citep{ghassemi2018you} recordings~(Sleep-PSG), and explore post-training TimeRLM. On average, TimeRLM improves upon all evaluated \ac{TSLM} baselines on all evaluated benchmarks, and transfers zero-shot to unseen clinical domains where post-training on synthetic data retains the model's capabilities. We release all code and data for both TimeRLM and AnomalyXL fully open-source.

\begin{figure*}[t]
\centering
\includegraphics[width=0.98\textwidth]{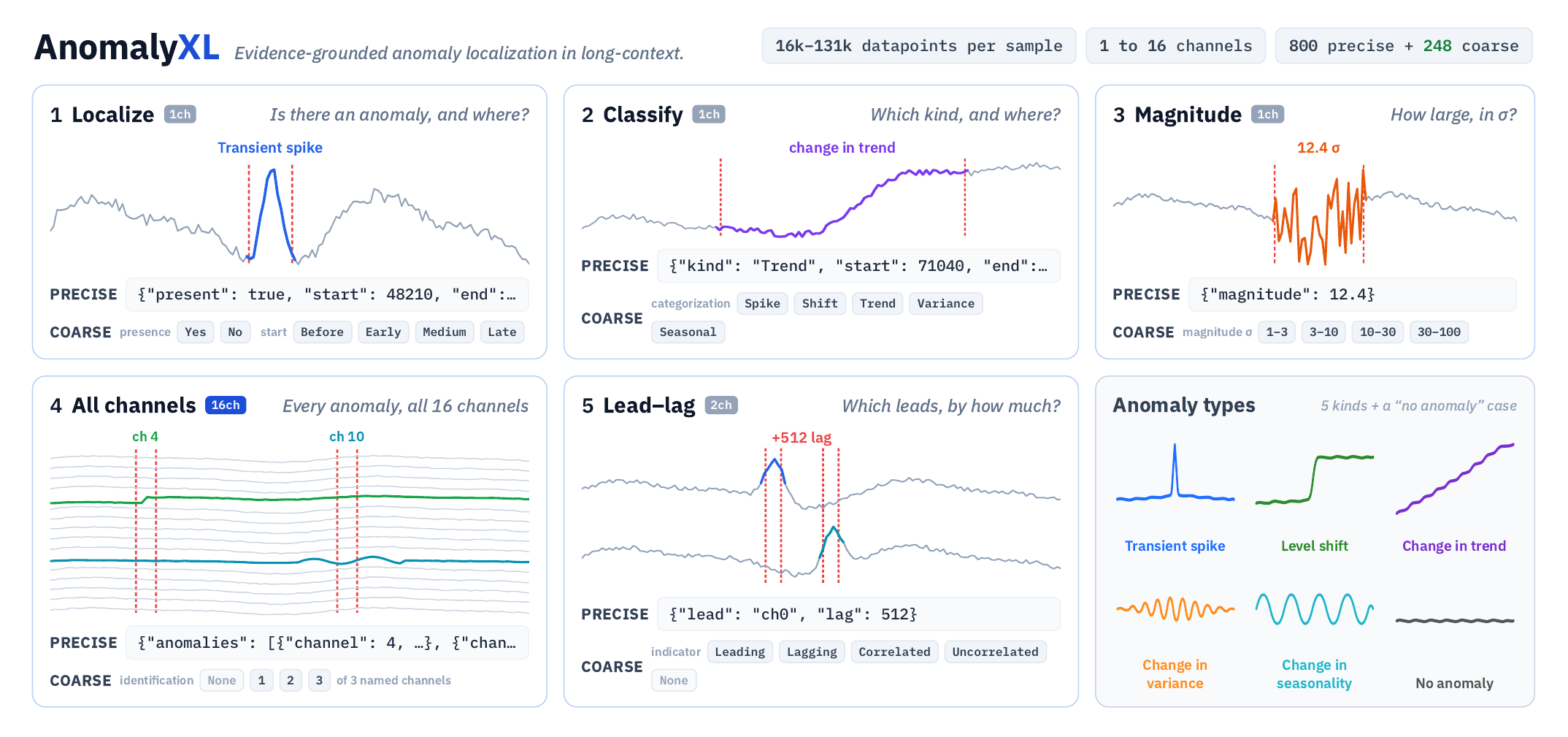}
\caption{\textbf{AnomalyXL: evidence-grounded anomaly localization.}
Five task families (localize, classify, measure magnitude, localize in all channels, lead-lag)
in two formulations over the \emph{same} signal: AnomalyXL-MCQ and AnomalyXL-Localize.}
\label{fig:anomalyxl}
\end{figure*}


\section{Related Work}
\label{sec:relatedwork}

In the following we explore related work to TimeRLM among TSLMs, anomaly detection benchmarks, and agentic methods for long context retrieval.

\paragraph{Time-Series Language and Foundation Models.}
\acp{TSLM} are multimodal \acp{LLM} that incorporate time-series as a native
modality~\citep{langer2026opentslm}. \acp{TSLM} differ in  
how each model processes time-series of length $L$.
\emph{Full-resolution adapters} pass raw tokens or $N\,{=}\,L/P$ patch embeddings directly
to the \ac{LLM}, preserving signal fidelity but leading to quadratic compute and memory growth due to $\mathcal{O}(N^2)$ cost of self-attention~\citep{wang2024chattimeunifiedmultimodaltime, Xie_2025, vaswani2017attention}.
A different approach based on \emph{fixed-latent compression} places a module between the encoder and the \ac{LLM} that distills the signal into a constant number of tokens. ITFormer's summary tokens~\citep{wang2025itformer} and OpenTSLM-Flamingo's Perceiver Resampler~\citep{langer2026opentslm,jaegle2021perceiver} allow computational cost to remain almost constant at the \ac{LLM} level; however the compression ratio grows with longer time-series, compressing more data into the same number of fixed latent tokens.
A more recent line of work pairs inputs from a frozen time-series foundation model and rendered time-series plots with a \ac{VLM} backbone, using both modality tokens to reason over the time-series signal~\citep{xie2026arfbench}.

\paragraph{Anomaly Detection and Short-Context Prediction.}
Time-series anomaly detection is traditionally formulated as identifying anomalous points or subsequences, often by assigning anomaly scores to sliding-window segments~\citep{boniol2024divetimeseriesanomalydetection}.
With the advent of \acp{LLM}, a new paradigm of reasoning over signals in natural language to explain why specific segments are anomalous has emerged~\cite{Su2024LargeLMAnomaly}.
Benchmarks in this field typically evaluate anomaly detection accuracy over short, pre-segmented recordings, using semantic labels~\citep{oh2023ecg} and \ac{MCQ} answering formats~\citep{xie2026arfbench}.

\paragraph{Recursive Language Models and Long-Context Agents.}
Agentic frameworks decompose complex tasks into a series of \ac{LLM} reasoning steps and actions, leveraging tool use~\citep{yao2023react} and iterative retrieval~\citep{wang2024videoagent}.
\acp{RLM} take this concept further by treating context as an environment variable that a \textit{root} \ac{LLM} queries through code and which can recursively spawn \emph{sub-RLMs} on segments of this context, rather than ingesting it directly~\citep{zhang2026recursivelanguagemodels}.
In time-series, recent work shows how \acp{LLM} are capable of time-series analysis using code~\cite{xu2026multimodalmodelsreasonecg}. TimeSeriesScientist and ARTS orchestrate specialized forecasting and classification tools respectively, decoupling task-specific proficiency from global reasoning~\citep{zhao2025timeseriesscientistgeneralpurposeaiagent, zumarraga2026tshaystack}. ARTIST learns segment selection and time-series reasoning with reinforcement learning~\citep{messica2026artist}.


\section{Methods}
\label{sec:methods-rlm}

In the following, we present details on TimeRLM in Section~\ref{sec:methods:timerlm} and AnomalyXL in Section~\ref{sec:methods:anomalyxl}. TimeRLM is an \ac{RLM} formulation that enables reasoning over uncompressed multivariate time-series using code, whereas AnomalyXL is a synthetic long-context time-series benchmark. We additionally introduce a generalization set, sampled from real ECG and Sleep clinical recordings and explore post-training for the \ac{RLM} formulation.

\subsection{TimeRLM Formulation}
\label{sec:methods:timerlm}
We formulate TimeRLM as a Recursive Language Model~\citep{zhang2026recursivelanguagemodels} where the time-series is queryable from a context file.
We start with a text-only \ac{RLM}. Over multiple \textbf{turns} (i.e., \ac{RLM} agent actions), the model can emit Python into a code sandbox (i.e., \ac{REPL}) with access to standard python data science libraries, receiving each execution's \ac{REPL} output in its context. A JSON object containing the time-series is pre-defined as a variable. This setup allows the model to operate over the full recording directly through code (Fig.~\ref{fig:timerlm}).
We also explore a second \ac{RLM} setup in which the sandbox also provides access to python vision libraries, allowing the model to render time-series as a plot and perceive it visually.
In a third setup, we allow the model to recursively call instances of itself (\emph{sub-RLMs}) on the whole or a fraction of the time-series.

\paragraph{Anomaly Localization as a Decision Process.}
We formalize precise anomaly localization as a finite-horizon sequential decision process. Given
a recording $\mathbf{X}\in\mathbb{R}^{C\times L}$ over $C$ channels of length $L$ and a
question $q$, an LLM denoted as policy $\pi_\theta$ (the root \ac{LLM}) produces an answer $\hat{y}$
over a horizon of $N$ interactions ($t=1,\dots,N$).
At step $t$ the state $s_t=(\mathbf{X},q,h_t)$ pairs the recording with the
interaction history $h_t=(o_0,a_1,o_1,\dots,a_{t-1},o_{t-1})$, where $o_0$ is the initial observation the
environment exposes before any action is taken (the question, and, optionally, the recording itself) and the policy selects an action from
\[
  \mathcal{A}=\mathcal{C}\;\cup\;\{\,\mathtt{FINAL}(\hat{y})\,\},
\]
either executing code $c\in\mathcal{C}$ in the sandbox or committing a structured answer
that terminates the episode.
Executing code returns an observation $o_t=\mathrm{exec}(c_t,s_t)$ for $t\ge1$ appended to the history, and the episode terminates with a single verifiable reward
$r(\hat{y},y)\in[0,1]$. 
The Single-pass prediction is the special case $N=1$ in which $\mathcal{A}=\{\mathtt{FINAL}(\hat{y})\}$ and
$o_0$ already contains the signal, so the policy responds
directly from this observation, differing from
the recursive case in both the action space and the content of the initial observation.

\subsection{The AnomalyXL Benchmark}
\label{sec:methods:anomalyxl}

AnomalyXL is a long-context, multi-channel benchmark for evidence-grounded anomaly
localization: every answer requires precise retrieval of a time span, channel, or magnitude
(Fig.~\ref{fig:anomalyxl}). We generate samples from a stochastic
noise process with optional periodic/trend components, z-normalized per channel
($x \leftarrow (x-\mu_x)/\sigma_x$) so injected anomalies keep consistent scale across
heterogeneous base signals. Contexts span 16k--131k points and up to 16 channels.

\paragraph{Two formulations.}
\label{sec:methods-tasks-scoring}
Generated samples are posed in two answer formats. \textbf{AnomalyXL-MCQ} asks each
question against a fixed label set and is scored by $F_1$ over six categories:
presence, onset before or after a given timestamp, type, magnitude range, affected
channel, and lead versus lag (\S\ref{app:anomalyxl:metrics:coarse_rows}).
\textbf{AnomalyXL-Localize} instead asks for a structured JSON summary of every
detected anomaly (see Appendix \ref{sec:appendix:anomalyxl:json}), scored by task-specific
continuous metrics in $[0,1]$ across five tasks:

\begin{itemize}[leftmargin=1em,itemsep=1pt,topsep=2pt]
  \item \textbf{Localize} (1 channel) --- detect the presence of an anomaly and return the window, scored
        by \ac{IOU}~(Appendix \ref{app:anomalyxl:metrics:localize}).
  \item \textbf{Classify} (1) --- return the anomaly type, scored
        $\mathbf{1}\{\hat{k}=k\}\cdot\mathrm{IoU}$, so a wrong type scores zero  (Appendix \ref{app:anomalyxl:metrics:classify_with_evidence}).
  \item \textbf{Magnitude} (1) --- report $m$, with credit decaying linearly in relative
        error and reaching zero at $50\%$
        (Appendix \ref{app:anomalyxl:metrics:measure_magnitude}).
  \item \textbf{All channels} (16) --- return every anomalous channel and window, scored
        by set-matched $F_1$ over events, where a prediction matches when both the
        channel and an overlapping window are correct
        (Appendix \ref{app:anomalyxl:metrics:all_channels}).
  \item \textbf{Lead--lag} (2) --- return the direction and lag, with credit only for
        the correct direction, then decaying with relative error in the reported offset
        (Appendix \ref{app:anomalyxl:metrics:lead_lag}).
\end{itemize}

Temporal-overlap calculations (see \ref{app:anomalyxl:metrics:temporal_overlap}),
aggregation procedures, and the task--metric correspondence
(Table~\ref{app:anomalyxl:metrics:aggregation}) are given in
Appendix~\ref{sec:appendix-anomalyxl}. Anomalies are injected by applying a localized transformation to a window $x_{s:e}$,
scaled by a magnitude $m$ (in normalized channel units) drawn uniformly from $[2,6]$,
or log-uniformly from $[1,100]$ for the magnitude task. A \emph{level shift} adds a
signed offset $\pm m$ over $x_{s:e}$; a \emph{transient spike} adds a triangular pulse
peaking at $\pm m$; a \emph{seasonality} anomaly adds a sinusoid of amplitude $m$; a
\emph{variance} anomaly adds noise $\epsilon_t \sim \mathcal{N}(0,m^2)$; and a
\emph{trend} anomaly adds a ramp from $0$ to $\pm m$ that then persists. Localize and
all-channels include first-class no-anomaly cases (19.5\% and 23\% of rows; 7.8\% of
the test split).


\section{Experiments}
In the following, we describe our experiments for evaluating TimeRLM. In Section~\ref{sec:experiments:anomalyxl}, we evaluate TimeRLM on both AnomalyXL-MCQ and AnomalyXL-Localize, compare it against \ac{TSLM} and classical baselines, and describe an experiment for improving performance using reinforcement learning. Section~\ref{sec:experiments:arfbench} describes the evaluation on ARFBench, while Section~\ref{sec:experiments:unseen_holdout} shows how we assess zero-shot performance on an unseen holdout set from the clinical domain.

\subsection{Evaluation on AnomalyXL}
\label{sec:experiments:anomalyxl}
We evaluate four inference-time configurations on AnomalyXL: single-pass \acp{LLM} and TimeRLM, each in both text and image variants. Unless explicitly stated, all are evaluated without any training, using Qwen3.5-4B~\citep{qwenteam2026qwen35omnitechnicalreport}, GPT-4.1-mini~\citep{openai2023gpt4} and GPT-5.5~\citep{singh2025openaigpt5card} as progressively more powerful backbones. 
We disable sub-agent recursion for \acp{RLM} by default unless explicitly mentioned (marked as ``depth=1'').
The single-pass text variant serializes raw values into the prompt, uniformly subsampling at a fixed stride on long recordings when needed to fit the context window, following prior time-series question-answering protocols~\citep{xie2026arfbench}. For the image variant, the single-pass formulation receives a rendered plot of the signal instead of the serialized raw values~(see Section~\ref{sec:methods:timerlm} for details).

\paragraph{Baselines}
We compare against three purpose-built \acp{TSLM} using the same Qwen3.5-4B backbone: ChatTS ~\citep{Xie_2025}~(full resolution text), OpenTSLM-Flamingo~\citep{langer2026opentslm}~(compressed latents), and ITFormer~\citep{wang2025itformer}~(compressed tokens). We instantiate all three architectures with a frozen pretrained Chronos-2 time-series encoder~\citep{ansari2025chronos2} and a LoRA-adapted Qwen3.5-4B \ac{LLM} backbone~\citep{hu2022lora}. We further benchmark \emph{Toto-1.0-QA-Experimental}, a \ac{TSFM}--\ac{VLM} hybrid introduced with ARFBench~\citep{xie2026arfbench,cohen2025time}, which processes both the time-series and pre-rendered plots. It uses a Qwen3-VL-32B \ac{LLM} backbone and a frozen Toto time-series encoder~\citep{cohen2025time}. We fine-tune separate variants for AnomalyXL-MCQ and AnomalyXL-Localize, allowing each adapter to learn the corresponding answer format. Each variant is trained on $\approx 5{,}000$ generated examples for $5$ epochs, with early stopping based on validation loss. We also evaluate Toto without SFT~(denoted 'No SFT') using the open-source checkpoint, as it was trained for anomaly localization~\citep{xie2026arfbench}. All supervised baselines use the same LoRA configuration, with hyperparameter details provided in Appendix~\ref{sec:appendix-tslm-training}.
We also include a classical baseline with no language model on the same AnomalyXL train/validation splits used for the \acp{TSLM}. We pair standard anomaly-detection features (robust-$z$ exceedance geometry, half-series level and variance contrasts, rolling-variance and spectral change, PELT change points~\citep{killick2012optimal}, and matrix-profile discords~\citep{yeh2016matrix}) with one gradient-boosted head each for AnomalyXL-MCQ and AnomalyXL-Localize~\citep{chen2016xgboost} (Appendix~\ref{sec:appendix-classical}).

\paragraph{Improving performance via Reinforcement Learning on AnomalyXL.}
Since detecting and localizing anomalies yields verifiable rewards, we
investigate whether reinforcement learning can further improve performance. We post-train
TimeRLM with a Qwen3.5-4B \ac{RLM} backbone using \ac{GRPO}~\citep{deepseek-math} on
AnomalyXL-Localize, with full-parameter tuning and no supervised warm start. Rather than
relying on a fixed training set, we integrate the generator directly into the training
loop: every step synthesizes a fresh batch of recordings, so no sample repeats within a run. Formally, for a task with evidence label $y$, each rollout $i$ commits an answer $\hat{y}_i$
and receives a score $s_i = r(\hat{y}_i,y)\in[0,1],$
where $r$ is the outcome of the decision process, computed with the same AnomalyXL-Localize
parser and per-category metric. We train on the penalized rollout reward
$$R_i = s_i - \lambda_{\mathrm{tok}}\, s_i\, \kappa_i - \lambda_{\mathrm{fail}}\,\mathbf{1}\{\text{forced final}_i\},$$
where $\kappa_i\in[0,1]$ is the group-normalized token cost of rollout $i$ and
$\lambda_{\mathrm{tok}},\lambda_{\mathrm{fail}}$ weight the two cost penalties. Letting $T_i$
denote the number of tokens generated by rollout $i$ (across all turns of the episode,
excluding sandbox observations), we normalize within the group of $G$ rollouts on
the same task,
$$\kappa_i = \frac{T_i}{\max_{j\le G} T_j},$$
so that cost is measured relative to the most verbose rollout of the same task rather than
against an absolute budget, leaving exploration on harder tasks unpenalized. For the same
reason, the token-cost term applies only to rollouts with nonzero task reward ($s_i>0$).
For each sampled task we draw a group of $G$ rollouts and compute Dr.~GRPO advantages from
the penalized rewards,
$$A_i = R_i - \frac{1}{G}\sum_{j=1}^{G} R_j$$
without group-standard-deviation normalization~\citep{liu2025understanding}. Groups with
zero reward variance yield $A_i=0$ for every rollout and are dropped~\citep{yu2026dapo}.

We train for $300$ steps with group size $G=8$, batch size $128$, and a 64k-token training
sequence length. Rollouts follow the TimeRLM text variant environment with a budget of 15
agent turns, a token-cost coefficient $\lambda_{\mathrm{tok}}=0.1$, and a flat forced-final
penalty $\lambda_{\mathrm{fail}}=0.25$. We report final checkpoint performances. Anomaly-free cases are
withheld for the first 220 steps and reintroduced thereafter, preventing early rollouts from
over-using the ``no anomaly'' answer before the policy has reinforced anomaly-localization
behavior.

\subsection{Evaluation on ARFBench Tier 2}
\label{sec:experiments:arfbench}

Beyond AnomalyXL, we evaluate TimeRLM on the Tier 2 tasks of ARFBench~\citep{xie2026arfbench}. This includes 306 questions that ask for the properties of an anomaly~(identification, start, end, magnitude, and categorization) in multiple-choice form over real telemetry data, comparable to AnomalyXL-MCQ.  We instantiate TimeRLM using Qwen3.5-4B zero-shot with and without \ac{RL} on AnomalyXL-Localize and GPT-5.5 on its text and image variants. As baselines, we use GPT-5.5 in the same text and image variants as detailed in Section~\ref{sec:experiments:anomalyxl}.  Additionally, we explore Toto-1.0-Qwen3 32B as TSLM and Toto-1.0-QA-Experimental 32B as TSVLM, one reported from the paper's results as no checkpoint is provided and the other using the checkpoint provided by the authors~\cite{xie2026arfbench}.

\subsection{Evaluating generalization to unseen data}
\label{sec:experiments:unseen_holdout}

To evaluate generalization to unseen tasks, we construct holdout test sets from clinical domains by sampling 250 windows from the Long-Term Atrial Fibrillation (LTAF) \ac{ECG} dataset~\citep{petrutiu2007ltaf} and 250 from the Sleep Polysomnography (PSG) corpus~\citep{ghassemi2018you}. The LTAF windows contain two channels, each with 7.7k--130k samples, while the Sleep-PSG windows contain four respiratory channels, each with 12k--128k samples. We formulate each window as a ``classify-with-evidence'' task, following AnomalyXL-Localize, and report macro-F1 for classification and intersection over union (IoU) for localization (Appendix~\ref{sec:appendix-generalization}). Recordings are derived from windows of one to seventeen minutes, each containing at most one annotated anomaly, such as an ectopic beat in an \ac{ECG} signal or a respiratory arousal in a sleep recording. Windows are selected so that no neighboring annotated anomaly is present, while 20\% of the windows are anomaly-free to provide challenging negative examples. The median anomaly occupies 0.5\% of an LTAF window and 5.7\% of a Sleep-PSG window. As baselines, we evaluate the same TSLMs used for AnomalyXL~(Section~\ref{sec:experiments:anomalyxl}). None of the models are trained on these datasets.

\subsection{Results}
\begin{table*}[h!t]
\centering
\scriptsize
\scriptsize
\setlength{\tabcolsep}{1mm} 
\renewcommand{\arraystretch}{1.0}
\providecommand{\hd}[1]{{\scriptsize #1}}
\definecolor{hmcol}{HTML}{2166AC}   
\providecommand{\hm}[2]{\cellcolor{hmcol!#1}#2}
\begin{tabularx}{\textwidth}{@{}>{\scriptsize\itshape\raggedright\arraybackslash}p{12mm}@{\hspace{4pt}}>{\scriptsize\raggedright\arraybackslash}p{32mm}NNNNNN!{\vrule width 0.4pt}NBNNN@{}}
\toprule
 & \textbf{Model} & \multicolumn{6}{c}{\textbf{AnomalyXL-MCQ}} & \multicolumn{5}{c}{\textbf{AnomalyXL-Localize}} \\
\cmidrule(lr){3-8} \cmidrule(lr){9-13}
 & & \hd{Pres(F1)} & \hd{ST(F1)} & \hd{Cat(F1)} & \hd{Mag(F1)} & \hd{Id(F1)} & \hd{Ind(F1)} & \hd{Loc(IoU)} & \hd{C+E(cls$\cdot$IoU)} & \hd{MMag(err)} & \hd{LAC(F1)} & \hd{LL(cls$\cdot$err)} \\
Samples   & & \hd{(40)} & \hd{(40)} & \hd{(42)} & \hd{(42)} & \hd{(42)} & \hd{(42)} & \hd{(200)} & \hd{(200)} & \hd{(200)} & \hd{(100)} & \hd{(100)} \\
\midrule

\multirow{1}{=}{Classical}
 & Features + GBT heads & \hm{47}{0.861} & \hm{54}{\textbf{0.981}} & \hm{35}{\underline{0.643}} & \hm{52}{\textbf{0.945}} & \hm{41}{0.739} & \hm{30}{0.521} & \hm{9}{0.160} & \hm{1}{0.022} & \hm{38}{\underline{0.685}} & \hm{30}{\underline{0.549}} & \hm{5}{0.108} \\
\midrule

\multirow{5}{=}{TSLM}
 & ChatTS Qwen3.5-4B & \hm{26}{0.467} & \hm{8}{0.144} & \hm{8}{0.137} & \hm{6}{0.118} & \acs{OOM}\textsuperscript{*}  & \hm{13}{0.175} & \hm{0}{0.003} & \hm{0}{0.000} & \hm{5}{0.092} & \acs{OOM}\textsuperscript{*} & \hm{16}{0.350} \\
 & OpenTSLM Qwen3.5-4B & \hm{26}{0.467} & \hm{7}{0.127} & \hm{6}{0.101} & \hm{6}{0.118} & \hm{1}{0.010} & \hm{12}{0.167} & \hm{0}{0.008} & \hm{0}{0.000} & \hm{4}{0.079} & \hm{13}{0.230} & \hm{19}{0.350} \\
 & ITFormer Qwen3.5-4B & \hm{26}{0.467} & \hm{1}{0.014} & \hm{8}{0.146} & \hm{10}{0.183} & \hm{18}{0.333} & \hm{9}{0.254} & \hm{0}{0.004} & \hm{0}{0.001} & \hm{4}{0.078} & \hm{13}{0.230} & \hm{14}{0.333} \\
 & Toto Qwen3-VL 32B (no SFT) & \hm{44}{0.804} & \hm{33}{0.591} & \hm{12}{0.226} & \hm{20}{0.362} & \hm{5}{0.098} & \hm{22}{0.391} & \hm{1}{0.026} & \hm{0}{0.001} & \hm{16}{0.285} & \hm{12}{0.220} & \hm{18}{0.328} \\
 & Toto Qwen3-VL 32B & \hm{45}{0.812} & \hm{51}{0.921} & \hm{40}{\textbf{0.728}} & \hm{52}{\underline{0.944}} & \hm{17}{0.303} & \hm{20}{0.367} & \hm{18}{0.323} & \hm{4}{0.072} & \hm{45}{\textbf{0.822}} & \hm{12}{0.224} & \hm{31}{0.557} \\
\midrule

\multirow{6}{=}{Single-pass}
 & GPT-4.1-mini (text) & \hm{35}{0.640} & \hm{11}{0.194} & \hm{12}{0.215} & \hm{7}{0.120} & \hm{12}{0.219} & \hm{13}{0.167} & \hm{0}{0.005} & \hm{0}{0.000} & \hm{8}{0.154} & \hm{0}{0.000} & \hm{8}{0.139} \\
 & GPT-5.5 (text) & \hm{49}{0.886} & \hm{34}{0.620} & \hm{23}{0.420} & \hm{43}{0.774} & \hm{44}{0.803} & \hm{18}{0.476} & \hm{5}{0.096} & \hm{0}{0.001} & \hm{28}{0.506} & \hm{9}{0.170} & \hm{9}{0.212} \\
 & Qwen3.5-4B (text) & \hm{40}{0.729} & \hm{9}{0.163} & \hm{14}{0.258} & \hm{26}{0.476} & \hm{0}{0.000} & \hm{7}{0.199} & \hm{2}{0.035} & \hm{0}{0.001} & \hm{2}{0.034} & \hm{0}{0.000} & \hm{7}{0.122} \\
 & GPT-4.1-mini (image) & \hm{50}{\underline{0.902}} & \hm{38}{0.685} & \hm{10}{0.174} & \hm{8}{0.140} & \hm{8}{0.143} & \hm{14}{0.280} & \hm{8}{0.139} & \hm{0}{0.009} & \hm{11}{0.197} & \hm{6}{0.100} & \hm{11}{0.191} \\
 & GPT-5.5 (image) & \hm{55}{\textbf{1.000}} & \hm{51}{0.929} & \hm{10}{0.189} & \hm{37}{0.671} & \hm{29}{0.520} & \hm{42}{0.784} & \hm{18}{0.329} & \hm{2}{0.043} & \hm{24}{0.440} & \hm{7}{0.120} & \hm{35}{0.753} \\
 & Qwen3.5-4B (image) & \hm{49}{0.886} & \hm{49}{0.897} & \hm{11}{0.191} & \hm{7}{0.123} & \hm{14}{0.251} & \hm{33}{0.668} & \hm{6}{0.105} & \hm{1}{0.012} & \hm{3}{0.051} & \hm{4}{0.065} & \hm{16}{0.348} \\
\midrule

\multirow{5}{=}{\bfseries\upshape TimeRLM\\text\\(ours)}
 & GPT-4.1-mini & \hm{35}{0.631} & \hm{33}{0.594} & \hm{15}{0.270} & \hm{8}{0.154} & \hm{21}{0.374} & \hm{23}{0.529} & \hm{15}{0.273} & \hm{4}{0.075} & \hm{19}{0.352} & \hm{5}{0.088} & \hm{15}{0.291} \\
 & GPT-5.5 & \hm{35}{0.640} & \hm{50}{0.916} & \hm{31}{0.567} & \hm{44}{0.793} & \hm{46}{0.845} & \hm{52}{0.952} & \hm{37}{\underline{0.677}} & \hm{39}{\underline{0.717}} & \hm{27}{0.496} & \hm{25}{0.461} & \hm{53}{0.958} \\
 & GPT-5.5 (depth-1) & \hm{52}{0.937} & \hm{49}{0.895} & \hm{32}{0.585} & \hm{47}{0.846} & \hm{51}{\textbf{0.921}} & \hm{53}{0.976} & \hm{36}{0.656} & \hm{41}{\textbf{0.745}} & \hm{30}{0.554} & \hm{30}{\textbf{0.551}} & \hm{53}{\underline{0.978}} \\
 & Qwen3.5-4B & \hm{38}{0.687} & \hm{37}{0.673} & \hm{18}{0.325} & \hm{24}{0.445} & \hm{14}{0.253} & \hm{28}{0.684} & \hm{11}{0.205} & \hm{7}{0.119} & \hm{15}{0.272} & \hm{0}{0.003} & \hm{21}{0.398} \\
 & Qwen3.5-4B (RL) & \hm{41}{0.739} & \hm{39}{0.712} & \hm{14}{0.252} & \hm{28}{0.505} & \hm{19}{0.354} & \hm{44}{0.815} & \hm{30}{0.538} & \hm{20}{0.364} & \hm{28}{0.510} & \hm{13}{0.244} & \hm{33}{0.687} \\
\midrule

\multirow{3}{=}{\bfseries\upshape TimeRLM\\image\\(ours)}
 & GPT-4.1-mini & \hm{36}{0.657} & \hm{43}{0.785} & \hm{13}{0.240} & \hm{7}{0.133} & \hm{18}{0.330} & \hm{21}{0.553} & \hm{20}{0.369} & \hm{7}{0.126} & \hm{21}{0.389} & \hm{4}{0.076} & \hm{21}{0.350} \\
 & GPT-5.5 & \hm{42}{0.765} & \hm{51}{\underline{0.933}} & \hm{24}{0.432} & \hm{44}{0.801} & \hm{49}{\underline{0.886}} & \hm{52}{\textbf{0.976}} & \hm{38}{\textbf{0.682}} & \hm{38}{0.693} & \hm{26}{0.480} & \hm{30}{0.537} & \hm{53}{\textbf{0.988}} \\
 & Qwen3.5-4B & \hm{37}{0.664} & \hm{44}{0.793} & \hm{17}{0.307} & \hm{28}{0.505} & \hm{21}{0.381} & \hm{23}{0.675} & \hm{11}{0.208} & \hm{6}{0.109} & \hm{10}{0.176} & \hm{0}{0.004} & \hm{19}{0.345} \\
\bottomrule
\end{tabularx}
\caption{\textbf{AnomalyXL results.} For metrics see Table~\ref{tab:anomalyxl_metrics}. Number of samples listed below each task. \acp{LLM}/\acp{RLM} are zero-shot, \acp{TSLM} are fine-tuned.
\textbf{Bold}/\underline{underline} marks best/second-best.
\textit{MCQ:} Pres = presence, ST = start time, Cat = categorization,
Mag = magnitude, Id = identification, Ind = indicator.
\textit{Localize:} Loc = localize, C+E = classify + evidence,
MMag = measure magnitude, LAC = localize all channels, LL = lead/lag + magnitude.
\textit{*~OOM: Out of Memory.}}
\label{tab:main}
\end{table*}

In the following we highlight results and findings based on the experiments introduced in Sections~\ref{sec:experiments:anomalyxl} to~\ref{sec:experiments:unseen_holdout}.
\paragraph{AnomalyXL}
Table~\ref{tab:main} provides an overview of the results on AnomalyXL. On AnomalyXL-MCQ, TimeRLM with a GPT-5.5 backbone with images and GPT-5.5 with text and depth recursion lead on identification and indicator,
reaching $0.921$ and $0.976$, against $0.803$ and $0.784$ for the best single-pass
configuration. Across the remaining tasks spanning presence, start time, categorization, and magnitude, the fine-tuned 32B Toto
wins categorization ($0.728$), and single-pass GPT-5.5 (image) wins on the presence
($1.000$) task~(Table~\ref{tab:main}).
The three \acp{TSLM} (ChatTS, OpenTSLM, and ITFormer) all score $0.467$ on presence, and achieve $0.144$, $0.127$ and $0.014$ on start time. The Qwen3.5-4B single-pass image baseline reaches $0.897$
on start time. The image variant improves over its text counterpart on presence and start
time for all three backbones, while text remains better on magnitude ($0.476$ vs.\ $0.123$ for
Qwen3.5-4B).
On AnomalyXL-Localize the gap widens on the tasks scored with \ac{IOU}. TimeRLM (image, GPT-5.5) leads on localization with $0.682$ IoU, against $0.329$ for the best single-pass variant and $0.323$
for fine-tuned 32B Toto. On classify-with-evidence every baseline stays below $0.08$ while
TimeRLM reaches $0.745$. TimeRLM with recursion depth-1 gives small gains on tasks requiring precise segment localization (C+E $0.717$ to $0.745$, LAC $0.461$ to $0.551$) using an average of $1.14$ sub-RLM calls per task.
TimeRLM on a Qwen3.5-4B improves localization from $0.035$ to $0.205$ \ac{IOU} compared to the single-pass text variant, and to
$0.538$ after \ac{RL} post-training on AnomalyXL, above both the 32B Toto
($0.323$) and single-pass GPT-5.5 ($0.329$). Classify-with-evidence rises from $0.001$ to
$0.364$ for the same configuration.
On measure magnitude~(MMag), the task requiring a single scalar answer, Toto achieves $0.822$ and the classical
baseline $0.685$. The best TimeRLM variant based on GPT-5.5 with image modalities achieves ($0.551$).

\paragraph{ARFBench}
\begin{table}[h!t]
\centering
\scriptsize
\setlength{\tabcolsep}{3.5pt}
\begin{tabularx}{1.0\linewidth}{llXXXXXX}
\toprule
 & & Iden. & Start & End & Magn. & Categ. & Total \\
 & {\scriptsize \textit{n}} & {\scriptsize(38)} & {\scriptsize(56)} & {\scriptsize(32)} & {\scriptsize(76)} & {\scriptsize(104)} & {\scriptsize(306)} \\
\midrule
 & GPT-5.5 text  & 47.4 & 37.5 & \underline{65.6} & 48.7 & 58.7 & 51.6 \\
 & GPT-5.5 image & 34.2 & \textbf{51.8} & \textbf{71.9} & \underline{63.2} & \underline{66.3} & \underline{59.5} \\
 & Toto (TSLM)\textsuperscript{*}  & 10.5 & 35.7 & 34.4 & 47.4 & \textbf{71.2} & 47.4 \\
 & Toto (TSVLM)\textsuperscript{\dag} & 50.0 & 23.2 & 46.9 & 59.2 & 64.4 & 52.0 \\
\rowcolor{ourrow}
 & \emph{TimeRLM-text (Qwen3.5 4B)} & 31.6 & 30.4 & 46.9 & 36.8 & 31.7 & 34.3 \\
\rowcolor{ourrow}
 & \emph{TimeRLM-text (Qwen3.5 4B - RL)} & \textbf{65.8} & 26.8 & 40.6 & 48.7 & 33.7 & 40.8 \\
\rowcolor{ourrow}
 & \emph{TimeRLM-text (GPT 5.5)}  & 42.1 & \underline{50.0} & \underline{65.6} & 51.3 & 60.6 & 54.6 \\
\rowcolor{ourrow}
\multirow{-8}{*}{\rotatebox[origin=c]{90}{\textit{Accuracy}}}
 & \emph{TimeRLM-image (GPT 5.5)} & \underline{60.5} & \underline{50.0} & \underline{65.6} & \textbf{64.5} & 64.4 & \textbf{61.4} \\
\cmidrule[0.2pt](l{0pt}r{0pt}){2-8}
 & GPT-5.5 text  & 49.1 & 36.3 & 51.8 & 46.7 & 53.0 & 47.8 \\
 & GPT-5.5 image & 32.4 & \textbf{51.6} & \textbf{56.1} & 59.4 & \underline{64.0} & \underline{55.8} \\
 & Toto (TSLM)\textsuperscript{*}  & 17.5 & 41.3 & 23.0 & 35.9 & \textbf{66.2} & 43.6 \\
 & Toto (TSVLM)\textsuperscript{\dag} & 50.0 & 18.9 & 36.2 & 48.9 & 56.2 & 44.7 \\
\rowcolor{ourrow}
 & \emph{TimeRLM-text (Qwen3.5 4B)} & 44.5 & 42.7 & 41.3 & 46.2 & 24.0 & 37.3 \\
\rowcolor{ourrow}
 & \emph{TimeRLM-text (Qwen3.5 4B - RL)} & \textbf{74.5} & 41.1 & 42.2 & \textbf{68.3} & 29.0 & 48.0 \\
\rowcolor{ourrow}
 & \emph{TimeRLM-text (GPT 5.5)}  & 42.5 & \underline{49.4} & 51.6 & 44.3 & 53.4 & 48.9 \\
\rowcolor{ourrow}
\multirow{-8}{*}{\rotatebox[origin=c]{90}{\textit{Macro-F1}}}
 & \emph{TimeRLM-image (GPT 5.5)} & \underline{58.7} & \underline{49.4} & \underline{52.8} & \underline{60.3} & 61.9 & \textbf{57.9} \\
\midrule
\multicolumn{8}{@{}l}{\scriptsize \textsuperscript{*}\,Toto-1.0-Qwen3 32B (TSFM--LLM), numbers taken from the ARFBench paper} \\
\multicolumn{8}{@{}l}{\scriptsize \textsuperscript{\dag}\,Toto-1.0-QA-Experimental 32B (TSFM--VLM), reproduced (our local rerun)} \\
\bottomrule
\end{tabularx}

\caption{\textbf{ARFBench Tier 2 results}. Accuracy and Macro-F1 as defined in Table~\ref{tab:anomalyxl_metrics}.
\textbf{Bold}/\underline{underline}: best/second.}
\label{tab:arfbench}
\end{table}
On ARFBench Tier 2, TimeRLM leads both metrics: the image variant reaches $61.4\%$ accuracy and $57.9\%$ macro-F1, ahead of single-pass GPT-5.5 ($59.5\%$, $55.8\%$) and both Toto variants~(Table~\ref{tab:arfbench}).
This pattern is consistent with AnomalyXL-MCQ, where TimeRLM on a GPT-5.5 backbone leads among the \ac{LLM}-based methods, with Toto Qwen3-VL 32B (SFT) close in performance.

\vspace{-0.3em}

\paragraph{Generalization set (ECG + sleep).}
On zero-shot evaluation, TimeRLM leads both sets on classification and localization (Table~\ref{tab:generalization}).
TSLMs retain some ability to classify, but no \ac{TSLM} exceeds an IoU of $0.004$. This includes Toto, whose fine-tuned version outperforms some TimeRLM configurations on localization in AnomalyXL-Localize. 
TimeRLM with recursion spawns an average of $2.6$ sub-agents per task, compared to $1.14$ in AnomalyXL-Localize, and increases performance overall.
\begin{table}[h!t]
{\scriptsize
\begin{tabularx}{1.0\linewidth}{@{}>{\raggedright\arraybackslash\itshape}p{1.15cm} l XX!{\vrule width 0.4pt}XX@{}}
\toprule
 & \textbf{Model} & \multicolumn{2}{c}{\textbf{LTAF (ECG)}} & \multicolumn{2}{c}{\textbf{Sleep-PSG}} \\
 & & F1 & IoU & F1 & IoU \\
\midrule
& ChatTS                              & 0.000 & 0.000 & \acs{OOM}\textsuperscript{*} & \acs{OOM}\textsuperscript{*} \\
& OpenTSLM                            & 0.000 & 0.000 & 0.083 & 0.004 \\
& ITFormer                            & 0.331 & 0.000 & 0.330 & 0.000 \\
& Toto (No SFT)\textsuperscript{\dag} & 0.000 & 0.000 & 0.027 & 0.000 \\
\multirow{-5}{1.1cm}{\itshape TSLMs}
& Toto (SFT)\textsuperscript{\ddag}   & 0.000 & 0.000 & 0.000 & 0.000 \\
\midrule
\rowcolor{black!12} & GPT-4.1-mini      & 0.457 & 0.025 & 0.255 & 0.031 \\
\rowcolor{black!12} & GPT-5.5           & 0.724 & 0.165 & 0.262 & 0.147 \\
\rowcolor{black!12} & GPT-5.5 (depth-1) & \textbf{0.798} & \textbf{0.271} & \textbf{0.348} & \textbf{0.270} \\
\rowcolor{black!12} & Qwen3.5-4B        & 0.357 & 0.012 & 0.174 & 0.011 \\
\rowcolor{black!12}
\multirow{-5}{1.1cm}{\itshape TimeRLM-\\text (Ours)}
& Qwen3.5-4B (RL)   & 0.446 & 0.016 & \underline{0.340} & 0.002 \\
\midrule
\rowcolor{black!12} & GPT-4.1-mini & 0.414 & 0.052 & 0.201 & 0.063 \\
\rowcolor{black!12} & GPT-5.5      & \underline{0.780} & \underline{0.192} & 0.314 & \underline{0.220} \\
\rowcolor{black!12}
\multirow{-3}{1.1cm}{\itshape TimeRLM-\\image (Ours)}
& Qwen3.5-4B   & 0.278 & 0.009 & 0.168 & 0.004 \\
\midrule
\multicolumn{6}{@{}l}{\scriptsize \textsuperscript{*}\,Out of memory \textbar{} \textsuperscript{\dag}\,Toto-1.0-QA-Experimental 32B \textbar{} \textsuperscript{\ddag}\,SFT on AnomalyXL-Localize} \\
\bottomrule
\end{tabularx}
}
\captionof{table}{\textbf{Zero-shot results}. Performance on unseen LTAF and Sleep sets. \textbf{Bold}/\underline{underline}: best/second per column.}
\label{tab:generalization}
\vspace{-1em}
\end{table}
The TimeRLM variant trained using reinforcement learning on AnomalyXL maintains good performance with F1 rising from $0.36$ to $0.45$ on LTAF and from $0.17$ to $0.34$ on Sleep-PSG. However, performance on localization stays near zero~(Table~\ref{tab:generalization}).


\section{Discussion and Conclusion}

Our work investigated whether recursive LLMs that can query time-series data through code can improve performance on precise anomaly localization by (1) introducing a TimeRLM formulation, (2) introducing AnomalyXL as a new benchmark, and (3) evaluating performance across AnomalyXL, ARFBench, and an unseen holdout set. Our results support three readings: using RLMs is most advantageous on AnomalyXL-Localize and stems from the fact that the model is able to query the signal multiple times. The RLM approach boosts performance across both text and vision variants. Post-training a small open model on the RLM approach helps to maintain capabilities on zero-shot transfer to real out-of-domain signals.

\paragraph{Multiple choice questions limit retrieval measurement.}
Classical baselines, i.e., specialized models for anomaly detection, reach the highest weighted average ($0.81$) on AnomalyXL-MCQ, above every \ac{LLM}, \ac{TSLM}, and \ac{RLM} we evaluated. 
The same baseline models however, do not outperform TimeRLM under precise scoring, indicating that what they do not recover is the location of an anomaly itself.
These results show that MCQ tasks such as presence, start-time ranges, and magnitude bands are often solvable from aggregate signal statistics, allowing systems to score well without precise localization.
AnomalyXL-MCQ closely matches related works in anomaly localization tasks~\cite{cai2024timeseriesexam, xie2026arfbench}. AnomalyXL-MCQ is comparable to ARFBench tier 2~\cite{xie2026arfbench}, and we find that the relative performances of each model, i.e., Toto vs. TimeRLM, are comparable across both benchmarks indicating similarity.
These findings suggest that the multiple-choice format commonly utilized in existing \ac{TSLM} anomaly detection benchmarks over-reports and under-investigates the retrieval abilities of \ac{TSLM} anomaly localization architectures.

\paragraph{Recursion increases anomaly localization capabilities.}
On AnomalyXL-MCQ, frontier backbones perform strongly in the single-pass setting: GPT-5.5 image reaches $1.000$ on presence and $0.929$ on start time. TimeRLM redistributes performance across metrics, trading presence ($1.000$ to $0.765$) for gains in identification ($0.520$ to $0.886$) and categorization ($0.189$ to $0.432$). On AnomalyXL-Localize, however, TimeRLM variants improve on all five tasks~(Table~\ref{tab:main}).
The largest gains occur on tasks requiring citing specific segments, including classify-with-evidence ($0.043$ to $0.693$) and localization ($0.329$ to $0.682$). Measured magnitude, which requires a scalar answer, is the exception and remains nearly flat ($0.440$ to $0.480$ for image and $0.506$ to $0.496$ for text).
Rendering the signal as an image helps in the MCQ single-pass setting: for GPT-5.5, it raises start time from $0.620$ to $0.929$ and localization from $0.096$ to $0.329$. This advantage does not carry over to TimeRLM. Text and image variants score within $0.03$ \ac{IOU} on localization ($0.677$ vs.\ $0.682$) and within $0.03$ on classify-with-evidence ($0.717$ vs.\ $0.693$), while the image variant uses $3.1$ more agent turns on average.
TimeRLM performs best when allowed to spawn sub-agents that examine individual signal segments in depth, reaching $0.745$ on classify-with-evidence, $0.551$ on localize-all-channels, and $0.978$ on lead/lag, while recovering presence to $0.937$. These gains align with prior work on agentic search in other modalities~\citep{zhang2026recursivelanguagemodels, wang2024videoagent} and motivate further work on scaling inference-time search for \acp{TSLM} and using coding agents for long-context time-series retrieval.

\paragraph{On-policy learning and out-of-distribution transfer.}
Fine-tuned on AnomalyXL, \acp{TSLM} collapse on the unseen LTAF and Sleep-PSG generalization sets, with no model exceeding an \ac{IOU} of $0.004$. ITFormer reaches $0.331$ F1 on LTAF and $0.330$ on Sleep-PSG.
By contrast, reinforcement learning on the AnomalyXL generator improves TimeRLM with Qwen3.5-4B on clinical signal classification, AnomalyXL-MCQ and ARFBench. 
These results suggest that post-training on synthetic data preserves capabilities on unseen datasets while substantially improving out-of-domain transfer.
We hypothesize that reasoning in code provides a signal-agnostic learning environment: the action space remains invariant across signal distributions, allowing the model to learn general procedures for time-series anomaly localization. 
Related work similarly suggests that on-policy learning can reduce train--test mismatch and preserve prior capabilities better than standard off-policy \ac{SFT}~\citep{shenfeld2026selfdistillation}. For this reason, clean ablations separating code access and \ac{SFT}-matched post-training remain necessary.

\paragraph{Limitations and Future Work.}
To our knowledge, AnomalyXL is the first synthetic long-context anomaly benchmark released with its generator, enabling configurable difficulty. Its programmatic construction, however, limits the diversity of anomaly morphologies: injected events are parametric, whereas real anomalies may be ambiguous and context-dependent. AnomalyXL therefore primarily evaluates precise retrieval of numerical deviations rather than the full range of semantically meaningful anomalies. Although our generalization set partly addresses this limitation, future work should study long-context time-series localization on real-world data and beyond the clinical domains considered here. Direct comparisons between Toto and the other \acp{TSLM} should also be interpreted cautiously. Toto uses a time-series encoder pretrained on real observability data, is already post-trained for anomaly localization on real telemetry, and employs a 32B \ac{VLM}, whereas the general-purpose \acp{TSLM} share a 4B backbone.


\section*{Acknowledgments}
We thank Prime Intellect for providing open-source infrastructure for
RL environments and recursive language models, compute resources, and technical guidance that
supported this work.

\bibliography{references}

%
%
\clearpage
\appendix
\setcounter{secnumdepth}{2}


\section{TimeRLM Details}
\label{sec:appendix-harness}
This section further discusses the per task and context length performance of the best performing RLM configuration in AnomalyXL-Localize, GPT-5.5.
We additionally document the exact agent loop underlying TimeRLM during training and evaluation, Prime
Intellect's v1 \texttt{rlm-harness} protocol. The harness is a
persistent IPython agent run as a subprocess per rollout, instantiating the decision
process of the Methods section: a persistent code sandbox, a terminal structured answer, and a single
verifiable reward.

\begin{figure*}[t]
\centering
\includegraphics[width=\textwidth]{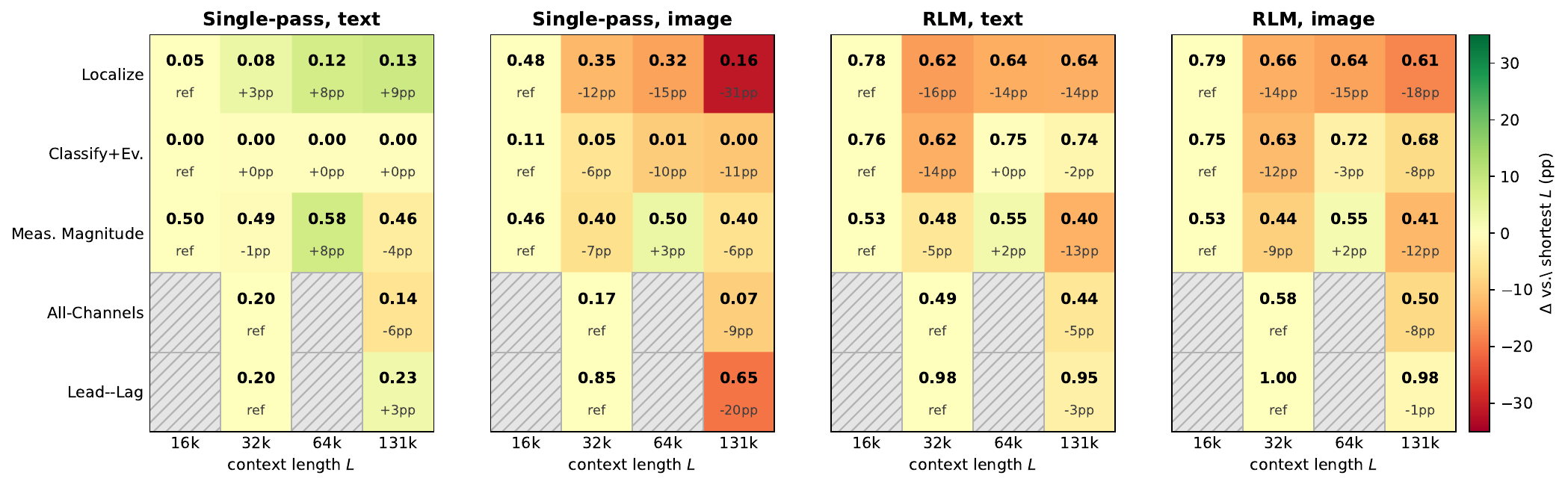}
\caption{\textbf{GPT-5.5 precise reward by category and context length}, for all four inference-time
configurations (verified against Table~\ref{tab:main} to three decimals). Each cell shows the mean reward
(bold) and its change in percentage points relative to that category's shortest sampled length;
color encodes this delta (green = improves with length, red = degrades).}
\label{fig:gpt55-heatmap}
\end{figure*}

\paragraph{Results by category and context length.}
The per-category averages in Table~\ref{tab:main} show that TimeRLM's precise gains concentrate on
evidence-pointing tasks, but they collapse away the length axis. Fig.~\ref{fig:gpt55-heatmap} breaks
GPT-5.5's precise score down by category \emph{and} context length for all four inference-time
configurations, and two patterns emerge that the averaged column hides. First,
classify-with-evidence is at or near zero for both single-pass variants at \emph{every} length (text:
$\le0.003$; image: $0.11\to0.00$ as $L$ grows), while the RLM holds $0.62$--$0.76$ at every length
regardless of modality. The collapse is uniform across the length range, not driven by the longest
contexts alone. Second, single-pass image localization degrades steadily with length, from $0.48$ at
$L=16$k to $0.16$ at $L=131$k, while the RLM's localize score stays within a narrow $0.61$--$0.79$ band
over the same range. With the exception for Measure Magnitude (a scalar answer not requiring precise retrieval scored with IoU) and the Localize task for single pass text, results indicate the relative single-pass/RLM gap widens with context length and degrades further in absolute terms from $L=32$k to $L=131$k, consistent with scanning 16 channels by eye or by serialized text becoming harder, not easier, as the series lengthens.

\begin{figure}[t]
\centering
\includegraphics[width=\linewidth]{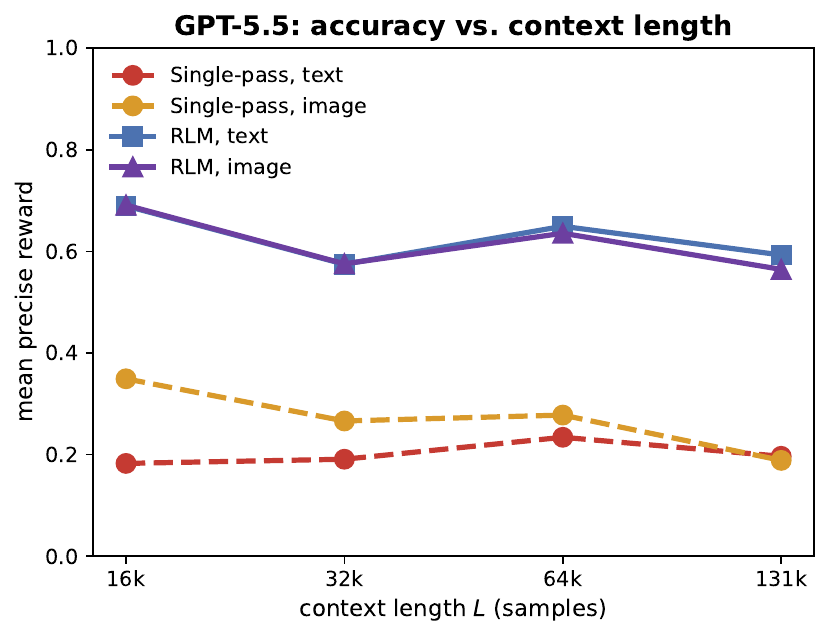}
\caption{\textbf{GPT-5.5 accuracy versus context length}, averaged over the three length-swept precise
categories (localize, classify-with-evidence, measure-magnitude) for each inference-time configuration.}
\label{fig:gpt55-acc-length}
\end{figure}

Fig.~\ref{fig:gpt55-acc-length} aggregates the three categories swept across all four lengths (localize,
classify-with-evidence, measure-magnitude) into a single accuracy-versus-length curve per configuration.
Averaged this way, single-pass image score drops by roughly $46\%$ from the shortest to the longest
context ($0.35\to0.19$), while single-pass text stays flat but low throughout ($0.18$--$0.23$); both RLM
curves remain within a tight $0.56$--$0.69$ band across the same length range, well above either
single-pass curve at every length and with no comparable widening or narrowing trend.

\paragraph{Prompts.}
The system prompt states a generic context, with task-specific instruction and answer options carried in the per-task prompt:

\begin{promptbox}[title=System prompt]
You are a specialized time-series agent that uses code to solve tasks. You solve tasks by breaking down
problems into sub-tasks, writing and executing code, observing results, and iterating one step at a time.
When you are done, stop calling tools and state your final answer.
\end{promptbox}

\begin{promptbox}[title=Precise prompt]
This is a critical application: when relevant, precisely locate the anomaly's start and end using
statistical thresholds --- be precise\ldots{} Analyze it with numpy and scipy in the IPython tool. When you
are done, stop calling the tool and state your final answer as a single JSON object matching the question's
schema exactly (e.g.\ \{"start": 12, "end": 87\}), with no prose and no markdown fences. Format failures
score zero.
\end{promptbox}

\paragraph{Turn budget.}
During RL, the episode is capped at 15 turns to maintain manageable memory budgets and speed during training; on the final turn, tools are withdrawn and a \emph{``FINAL TURN: your turn budget is exhausted and tools are disabled. State your final answer NOW\ldots{}''} notice is injected, including a $-0.25$ exhausted-budget penalty used at training time. The episode otherwise ends as soon as the model stops calling tools and states its final answer
directly.

\paragraph{Sandbox and packages.}
The agent executes in an isolated Python sandbox backed by a real Jupyter kernel, pre-seeded with
\texttt{numpy} and \texttt{scipy}. Tool output returned to the model is truncated at 8192 characters
per turn (\texttt{RLM\_MAX\_OUTPUT}).
Recursive \texttt{sub\_rlm()} calls are implemented but disabled (depth 0) throughout, as stated in
the Methods section. The answer is extracted by locating a balanced-brace JSON object containing the
schema's required keys, tolerating surrounding prose or markdown fences; a response from which no such
object can be recovered scores zero.

\section{Classical Baseline Details}
\label{sec:appendix-classical}

The classical baseline pairs a deterministic, untrained feature extractor with small
supervised task heads. No language model is involved at any stage.

\paragraph{Feature extraction.}
Series are decimated to at most 16{,}384 points by stride subsampling, and index-valued
predictions are mapped back through the stride. 25 features are computed per
channel:

\begin{itemize}
  \item \textbf{Robust-$z$ statistics.} Global median/MAD $z$-scores: $\max|z|$ and its
    position, the fraction of points beyond $4\sigma$ and $8\sigma$, exceedance-run
    geometry at $|z|>4$ (first and last exceedance, longest-run start, end and length,
    run count), and peak width at $|z|>6$.
  \item \textbf{Half-series contrasts.} Mean shift between halves (level shift), log
    variance ratio (variance change), and global linear slope (trend).
  \item \textbf{Rolling-variance localization.} Maximum and argmax of the robust-$z$ of a
    rolling standard deviation with window $L/64$, capturing variance-change onset.
  \item \textbf{Spectral change.} Dominant-frequency power ratio and frequency shift
    between halves, via the real FFT, capturing seasonality change.
  \item \textbf{Change points.} PELT with an $\ell_2$ cost on a 2{,}048-point grid:
    breakpoint count, first breakpoint, and the largest inter-segment mean jump with its
    location.
  \item \textbf{Matrix profile.} Maximum discord value and its location, on a
    4{,}096-point grid with window $L/64$.
  \item \textbf{Series length.} $\log_2$ of the raw, un-decimated recording length.
\end{itemize}

Each question is then described by 56 values: the worst channel's full feature vector (25),
the channel-mean of every feature (25), the channel count (1), and five cross-correlation
features between the two groups---best-lag correlation, lag as a fraction of length,
zero-lag correlation, and each group's own $\max|z|$---always present but zero-filled
outside the lead--lag and indicator tasks ($25+25+1+5=56$).

\paragraph{Task heads.}
Heads are gradient-boosted trees, one per question category (Table~\ref{tab:classical_heads}).
They are trained on \emph{exactly} the data used by the \ac{TSLM} baselines---the AnomalyXL
train and validation splits---with an independent set of heads fitted per regime, mirroring
the per-regime \ac{TSLM} and Toto fine-tunes; only the untrained, deterministic feature
extractor is shared between them.

\begin{table}[h]
\centering
\small
\begin{tabular}{@{}lll@{}}
\toprule
\textbf{Category} & \textbf{Head} & \textbf{Target} \\
\midrule
\multicolumn{3}{@{}l}{\textit{Coarse}} \\
Presence        & classifier & Yes / No \\
Start time      & classifier & start bin \\
Categorization  & classifier & anomaly kind \\
Magnitude       & classifier & $\sigma$ band \\
Identification  & per-channel binary & channel in gold set \\
Indicator       & classifier (paired) & 5 relations \\
\midrule
\multicolumn{3}{@{}l}{\textit{Precise}} \\
Localize        & presence clf.\ + span regr. & \texttt{start}, \texttt{end} \\
Classify + ev.  & kind clf.\ + span regr. & kind, \texttt{start}, \texttt{end} \\
Measure magn.   & regressor on $\log(1+\sigma)$ & \texttt{magnitude\_s} \\
Localize all ch.& per-channel clf. & anomaly list \\
Lead--lag       & direction clf.\ + lag regr. & direction, \texttt{lag} \\
\bottomrule
\end{tabular}
\caption{\textbf{One head per AnomalyXL category.} Span regressors predict start and end as
fractions of the recording length; intervals for localize-all-channels are read off the
longest positive run.}
\label{tab:classical_heads}
\end{table}

\paragraph{Decoding and evaluation.}
Head predictions are mapped deterministically into the benchmark's answer space: coarse heads
emit one of the row's option strings, and precise heads emit the structured JSON objects the
benchmark defines. Evaluation is a single pass over the same held-out test sets and the same
scorers used for every other row in Table~\ref{tab:main}.

Because the feature design encodes the task knowledge and the heads learn only thresholds and
combinations, this baseline receives, in effect, the information the language-model systems are
given in their prompts---the question and its category definition---but supplied to it by
construction rather than read from text.

\section{TSLM Fine-Tuning Configurations}
\label{sec:appendix-tslm-training}
In the following, we present details on the training configurations of all \ac{TSLM} baseline results displayed in the paper.
All finetuned \ac{TSLM} baselines share the same training configuration detailed below unless explicitly differentiated, and are separately finetuned on AnomalyXL-MCQ and AnomalyXL-Localize to allow each model variant to learn the answer schema required by each separate task, yielding a total of $8$ finetuning runs. An overview of the finetuned model variants is presented in Table~\ref{tab:tslm_training_runs}.

\begin{table*}[t]
\centering
\small
\setlength{\tabcolsep}{4pt}
\begin{tabular}{@{}lll@{}}
\toprule
\textbf{Model row} & \textbf{Signal interface} & \textbf{Fine-tuned regimes} \\
\midrule
ChatTS Qwen3.5-4B & full-resolution patch adapter & coarse, precise \\
Flamingo Qwen3.5-4B & Perceiver-Resampler fixed-latent compressor & coarse, precise \\
ITFormer Qwen3.5-4B & fixed summary-token aggregation & coarse, precise \\
Toto Qwen3-VL 32B & frozen Toto encoder with Qwen3-VL & coarse, precise \\
\bottomrule
\end{tabular}
\caption{\textbf{Supervised baseline fine-tuning runs.} The model names match the rows in Table~\ref{tab:main}. Each supervised baseline is trained once for the coarse answer format and once for the precise answer format.}
\label{tab:tslm_training_runs}
\end{table*}

ChatTS~\citep{Xie_2025}, OpenTSLM-Flamingo~\citep{langer2026opentslm} and ITFormer~\citep{wang2025itformer} use a frozen Chronos-2 time-series encoder~\cite{ansari2025chronos2} and a Qwen3.5-4B~\cite{qwenteam2026qwen35omnitechnicalreport} backbone adapted with LoRA~\cite{hu2022lora}.
Toto-1.0-QA-Experimental~\citep{xie2026arfbench} uses a pretrained Toto encoder~\citep{cohen2025time} attached to a Qwen3-VL-32B backbone, taken directly from the supplied ARFBench paper checkpoint in its base version, and further finetuned when mentioned, under the same conditions of all other \acp{TSLM}.

\paragraph{Hyperparameter configurations.}
All supervised fine-tunes use batch size 1, 5 epochs, an AdamW optimizer with learning rate $1e^{-4}$, weight decay 0.01, gradient clipping at norm 1.0, and a 0.03 warmup ratio. Early stopping monitors validation loss with patience 3. Validation is run after each epoch on 500 samples from each AnomalyXL-MCQ and AnomalyXL-Localize validation set, scored with the same AnomalyXL scoring rules used in the test set reported Table~\ref{tab:main}. The checkpoint achieving highest validation primary metric is selected as the evaluated baseline on the test set

\paragraph{LoRA adaptation.}
All fine-tuned language backbones are adapted with LoRA rather than full-model finetuning, set with rank 16, $\alpha=32$, dropout 0.05, no bias terms, and attention projection targets \texttt{q\_proj} and \texttt{v\_proj} while the time-series encoder remains frozen. For ChatTS, OpenTSLM-Flamingo, and ITFormer, LoRA is applied to the shared Qwen3.5-4B backbone and the architecture-specific time-series projection or fusion parameters are trainable. Toto follows the same optimization hyperparameters and LoRA settings, but applies them to its Qwen3-VL-32B text decoder. Its Toto encoder and projection modules remain frozen.

\paragraph{Data and scoring.}
The precise regime is trained on approximately 5{,}000 generated AnomalyXL examples using the same generator configuration as the evaluation set. Coarse models are finetuned to emit one of the multiple-choice options, and precise models to emit a structured JSON answer required by the task. The final reported numbers are computed with the benchmark's standard scorers: multiple-choice question accuracy for coarse tasks and the continuous evidence-grounded primary score for precise tasks.

\section{AnomalyXL Construction Details}
\label{sec:appendix-anomalyxl}

AnomalyXL signals are used both for precise, open-ended questions
answered in strict JSON, and for coarse multiple-choice
questions scored by choice accuracy. Table~\ref{tab:anomalyxl_tasks} gives the answer schema, metric,
and option set for each of the five tasks.

\begin{table*}[t]
\centering
\small
\begin{tabular}{@{}l m{0.36\textwidth} l m{0.30\textwidth}@{}}
\toprule
Task & Precise answer (JSON) & Metric & Coarse choices (MCQ) \\
\midrule
Localize     & \{"present": true, "start": 48210, "end": 48533\}              & IoU             & Presence: Yes / No;\quad Start Bin: Before / Early / Medium / Late \\
Classify     & \{"kind": "Trend", "start": 71040, "end": 73219\}              & kind$\times$IoU & Kind: Spike / Shift / Trend / Variance / Seasonal \\
Magnitude    & \{"magnitude\_sigma": 12.4\}                                   & rel.\ error     & Band: 1--3 / 3--10 / 10--30 / 30--100\,$\sigma$ \\
All channels & \{"anomalies": [\{"channel": 4, "start": 9000, "end": 9220\}, \dots]\} & set-matched F1 & Count: None / 1 / 2 / 3 of the named channels \\
Lead--lag    & \{"direction": "lead", "lag\_samples": 512\}                   & dir.+lag        & Indicator: Leading / Lagging / Correlated / Uncorrelated \\
\bottomrule
\end{tabular}
\caption{\textbf{The five AnomalyXL tasks, precise versus coarse.} Precise answers are strict JSON, scored
by the listed continuous metric in $[0,1]$; coarse answers are scored by choice accuracy. Magnitude's
relative error is converted to a score via $\max(0,\,1-\text{rel\_err}/0.5)$, a linear ramp from full
credit at an exact match to zero at $50\%$ relative error. The localize
no-anomaly case is answered with \{"present": false\}. See Fig.~\ref{fig:anomalyxl} for a schematic and
Fig.~\ref{fig:anomalyxl-examples} for worked examples drawn from the released data.}
\label{tab:anomalyxl_tasks}
\end{table*}

\subsection{Templated JSON response examples}
\label{sec:appendix:anomalyxl:json}
\begin{figure*}[t]
\centering
\includegraphics[width=\textwidth]{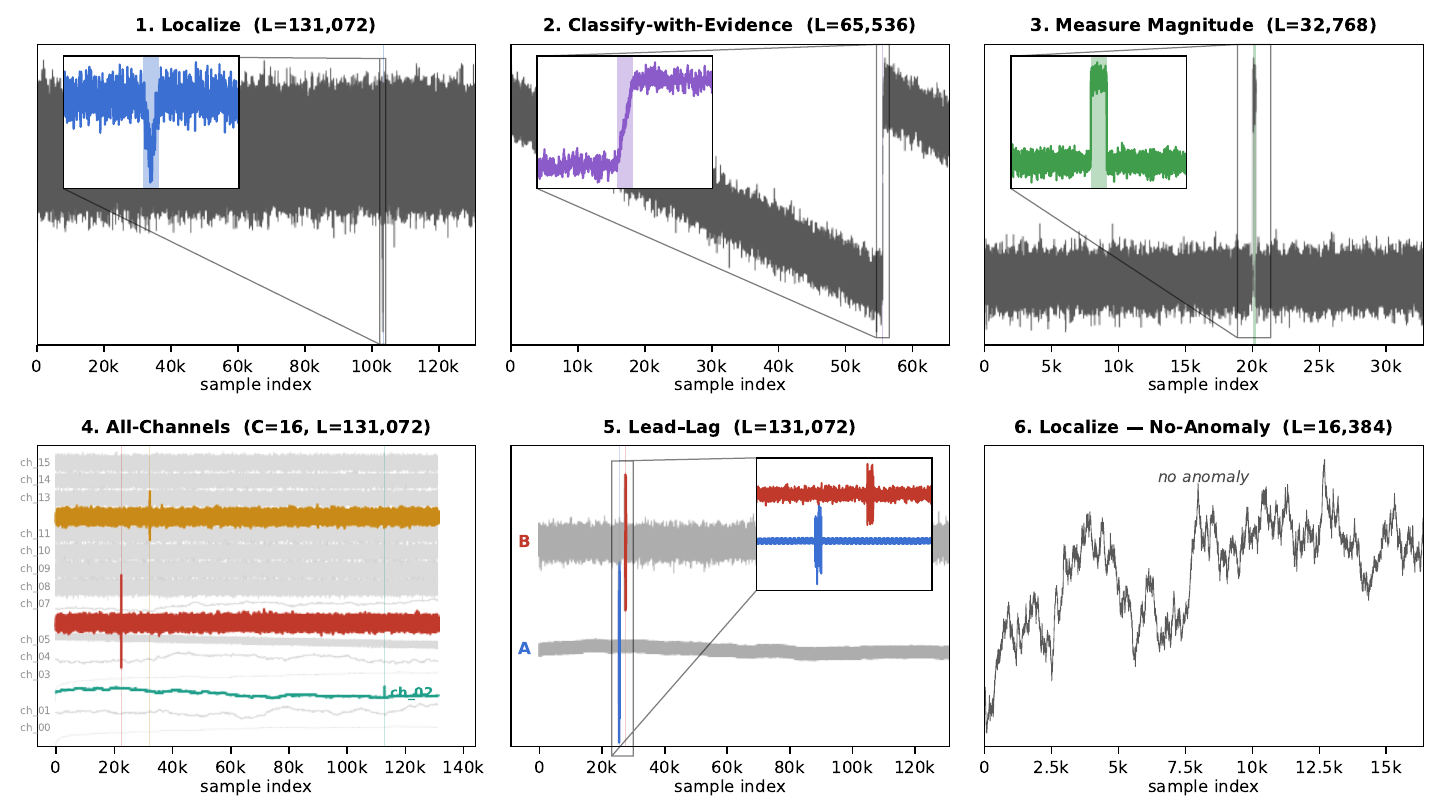}
\caption{\textbf{Six worked AnomalyXL-Localize examples}, sampled from the released dataset. Full-resolution
signal in gray with the gold window shaded; zoomed insets show the injected transform up close. Questions
and gold answers are given in the text above.}
\label{fig:anomalyxl-examples}
\end{figure*}

Fig.~\ref{fig:anomalyxl-examples} shows six sampled rows from the released \texttt{AnomalyXL-Localize} shard at full resolution (thin gray), with
the gold window shaded and, except for the no-anomaly case, a zoomed inset. Panel 1 (Localize,
$L=131{,}072$) asks whether an anomaly is present and where; gold \texttt{\{"present": true, "start":
103194, "end": 103352\}}. Panel 2 (Classify-with-Evidence, $L=65{,}536$) asks for the anomaly's kind and
window; gold \texttt{\{"kind": "Change in Trend", "start": 55469, "end": 55642\}}. Panel 3 (Measure
Magnitude, $L=32{,}768$) asks for the peak deviation in $\sigma$; gold \texttt{\{"magnitude\_sigma":
13.62\}}. Panel 4 (All-Channels, $C{=}16$, $L=131{,}072$) asks for every anomalous channel and its window;
gold anomalies fall on \texttt{ch\_02}, \texttt{ch\_06}, and \texttt{ch\_12} (widths 78--233 samples),
spread across the full series. Panel 5 (Lead--Lag, $L=131{,}072$) asks whether A leads or lags B and by
how many samples; gold \texttt{\{"direction": "lead", "lag\_samples": 2051\}}. Panel 6 (Localize,
no-anomaly case, $L=16{,}384$) is one of the first-class negatives; gold \texttt{\{"present": false\}}.
The exact question text for Classify-with-Evidence, from the released dataset, reads:

\begin{promptbox}[title={Classify-with-Evidence prompt (verbatim, $L=65{,}536$)}]
Identify the anomaly in the time series and its window. Respond with exactly one JSON object on a single
line, no other text: \{"kind": <label>, "start": <int>, "end": <int>\}. \texttt{kind} must be one of:
"Level Shift", "Transient Spike", "Change in Seasonality", "Change in Variance", "Change in Trend".
\texttt{start} is inclusive, \texttt{end} is exclusive. Each kind is defined as follows: a Level Shift is a
sustained step in mean\ldots{} (Here L = 65536.)
\end{promptbox}

\subsection{Scene synthesis.}
Every sample starts from a \emph{scene}: $C$ independent channels, each built as one base noise process
(white noise, a rescaled random-walk ``red'' noise, or an AR(1) process with $\phi=0.7$, drawn with equal
probability) plus an optional periodic component (probability 0.4; sine, square, or sawtooth with period
in $[16,256]$ samples) and an optional trend component (probability 0.3; linear, exponential, or
logarithmic drift, or a constant offset). Each channel is then z-normalized to unit std, so an anomaly
expressed in $\sigma$-units has a comparable visual scale regardless of which base processes were drawn.
Channels are anonymized (\texttt{ch\_00}, \texttt{ch\_01}, \dots, or \texttt{A\_}/\texttt{B\_}-prefixed for
the paired lead--lag task) so the model cannot exploit channel-name priors.

\subsection{Anomaly injection.}
An anomaly is one of five transforms applied to a contiguous window $[t_\text{start}, t_\text{end})$ of a
single target channel, \emph{after} z-normalization. A \emph{level shift} is a signed, sustained step of
$\text{magnitude}\times\sigma$). A \emph{transient spike} is a triangular pulse peaking at
$\text{magnitude}\times\sigma$ at the window's center, with zero at its edges. A \emph{change in seasonality} is a new sinusoid of period in $[4, \text{width}/2]$ samples, short enough that at least two cycles fit
inside the window, and added over the window). A \emph{change in variance} is extra Gaussian noise of std
$\text{magnitude}\times\sigma$ added sample-by-sample over the window, with a width floor raised to 64
samples so the change is a sustained regime rather than indistinguishable from a couple of spikes.
On the paired lead--lag task, this minimum is set to 11 samples. Finally, a \emph{change in trend} is 
a piecewise-linear ramp that reaches $\pm\text{magnitude}\times\sigma$ by the
window's end and then carries its terminal level forward, so it reads as a lasting slope change rather than
a transient pulse. Magnitude is drawn uniformly in $[2,6]\,\sigma$ for localize, classify, all-channels,
and lead--lag, and log-uniformly in $[1,100]\,\sigma$ for measure-magnitude, so that task alone stresses
both near-floor and extreme deviations. Window width is an \emph{absolute} sample count in $[8,256]$
regardless of $L$ --- not a fraction of the series --- which is the design choice that makes long context
hard: at $L=131{,}072$ an anomaly can occupy as little as $0.006\%$ of the series, indistinguishable from
noise in a full-resolution plot unless the reader zooms in, exactly as Fig.~\ref{fig:anomalyxl-examples}
shows.

\subsection{Channel topologies and sampling grid.}
\begin{figure}[t]
\centering
\includegraphics[width=\linewidth]{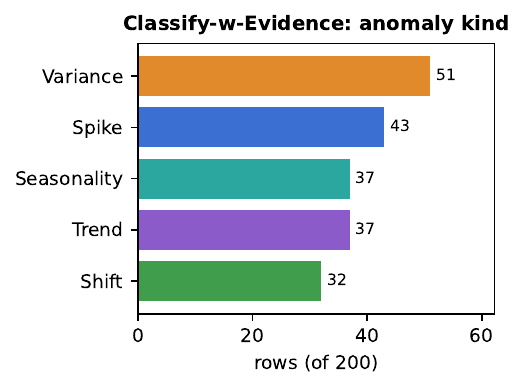}
\caption{\textbf{AnomalyXL-Localize dataset composition.} Anomaly-kind distribution within
classify-with-evidence.}
\label{fig:anomalyxl-distributions}
\end{figure}

Localize, classify-with-evidence, and measure-magnitude are single-channel ($C=1$) tasks swept over all
four context lengths, $L\in\{16384, 32768, 65536, 131072\}$. Localize-all-channels fixes $C=16$ and injects
$K\sim\mathcal{U}\{0,\dots,3\}$ anomalies on $K$ distinct channels with non-overlapping windows (channels
outside the injected set carry no anomaly), swept over $L\in\{32768, 131072\}$. Lead--lag-with-magnitude
splits $C=2$ channels into single-channel sub-scenes A and B and samples a temporal relation --- \emph{lead}
(A's anomaly precedes B's), \emph{lag} (A follows B), or \emph{independent} (disjoint, unrelated windows)
--- each with prior $1/3$; for lead/lag the offset is uniform in $[1, L/4]$ samples, so it never exceeds a
quarter of the series. Every $(L, \text{category})$ cell draws 50 rows from a seed derived deterministically, so extending the grid later does not reshuffle existing rows. Table~\ref{tab:anomalyxl_composition} summarizes the resulting 800-row precise shard. Fig.~\ref{fig:anomalyxl-distributions} plots the
classify-with-evidence anomaly-kind distribution.

\begin{table*}[t]
\centering
\small
\setlength{\tabcolsep}{5pt}
\begin{tabular}{@{}lcccl@{}}
\toprule
Category & $C$ & $L$ (samples) & Rows & Composition \\
\midrule
Localize            & 1  & 16k--131k & 200 & 161 present / 39 absent \\
Classify-w-Evidence  & 1  & 16k--131k & 200 & Variance\,51, Spike\,43, Seasonality\,37, Trend\,37, Shift\,32 \\
Measure Magnitude    & 1  & 16k--131k & 200 & $\sigma\in[1.0,99.8]$, median $12.6$ \\
All-Channels         & 16 & 32k, 131k & 100 & $K{=}0{:}23,\ 1{:}27,\ 2{:}25,\ 3{:}25$ \\
Lead--Lag            & 2  & 32k, 131k & 100 & lead\,33 / lag\,32 / independent\,35 \\
\midrule
Total (precise)      &    & 16k--131k & 800 & \\
\bottomrule
\end{tabular}
\caption{\textbf{AnomalyXL-Localize composition} (released \texttt{AnomalyXL-Localize} shard, seed 42).
Window widths are 8--256 samples throughout, independent of $L$; magnitude is in channel-$\sigma$ units.}
\label{tab:anomalyxl_composition}
\end{table*}

\subsection{Coarse projection.}
Rather than synthesizing a second, independent set of scenes, every coarse row is \emph{projected} from a
sampled precise row over the identical signal, with the source row's channels, injections, and coupling
copied, and only the question, options, and gold label rewritten into the corresponding MCQ
template (Table~\ref{tab:anomalyxl_coarse_projection}). This is what lets the paired precise/coarse design isolate the scoring axis, as the two regimes never differ in the underlying recording. The resulting shard has 248 rows.

\begin{table*}[t]
\centering
\small
\setlength{\tabcolsep}{5pt}
\begin{tabular}{@{}llcl@{}}
\toprule
Precise source & MCQ template(s) & Rows & Option set \\
\midrule
Localize             & Presence + Start-Time & 40$\times$2 & Yes/No; 4 quartile bins + No Anomaly \\
Classify-w-Evidence   & Categorization        & 42          & 4 of 5 kind labels (1 dropped at random) + No Anomaly \\
Measure Magnitude     & Magnitude             & 42          & 4 $\sigma$-bins + No Anomaly \\
All-Channels          & Identification        & 42          & 2 singletons + 1 pair + triplet + None, over 3 named channels \\
Lead--Lag             & Indicator             & 42          & 5 fixed options (Leading / Lagging / Correlated / Uncorrelated / No anomaly) \\
\midrule
Total (coarse)        &                        & 248         & \\
\bottomrule
\end{tabular}
\caption{\textbf{Coarse projection rules} (\texttt{AnomalyXL-MCQ}, projected from \texttt{AnomalyXL-Localize}
with seed 42). Localize yields two coarse rows per sampled precise row (presence and start-time); every
other category yields one.}
\label{tab:anomalyxl_coarse_projection}
\end{table*}


\section{AnomalyXL scoring and metrics}
\label{sec:appendix-anomalyxl-scoring}
\begin{table*}[t]
\centering
\footnotesize
\setlength{\tabcolsep}{4pt}
\renewcommand{\arraystretch}{1.25}
\begin{tabularx}{\textwidth}{@{}l >{\raggedright\arraybackslash}X >{\raggedright\arraybackslash}X l@{}}
\toprule
\textbf{Task} & \textbf{Required output, and quantity parsed from it} & \textbf{Score in $[0,1]$} & \textbf{Defined in} \\
\midrule
 
\multicolumn{4}{@{}l}{\textit{\textsc{AnomalyXL-Localize}} --- the first schema-valid JSON object in the final message is parsed; a row scores $0$ if none can be recovered} \\
\addlinespace[1pt]
 
Localize
& \texttt{\{"present": bool, "start": int, "end": int\}} $\rightarrow$ presence flag $\hat p$, and window $\hat w = [\hat s, \hat e)$ when $\hat p = 1$
& $1$ if $p = \hat p = 0$;\quad $\mathrm{IoU}(\hat w, w)$ if $p = \hat p = 1$;\quad $0$ on presence mismatch or missing window
& Section \ref{app:anomalyxl:metrics:localize}, Eq.~\eqref{eq:loc} \\
 
Classify
& \texttt{\{"kind": str, "start": int, "end": int\}} $\rightarrow$ label $\hat k$, matched by exact string equality against the five prompt options, and window $\hat w$
& $\mathbf{1}[\hat k = k]\cdot\mathrm{IoU}(\hat w, w)$; the label gates the window, so a wrong kind scores $0$ however well localized
& Section \ref{app:anomalyxl:metrics:classify_with_evidence}, Eq.~\eqref{eq:cls} \\
 
Magnitude
& \texttt{\{"magnitude\_sigma": float\}} $\rightarrow$ estimate $\hat m$, reduced to relative error $\rho = |\hat m - m|/m$; absent or non-numeric sets $\rho = 1$
& $\max(0,\, 1 - \rho/\tau)$ with $\tau = 0.5$: a linear ramp from full credit at an exact match to zero at $50\%$ relative error
& Section \ref{app:anomalyxl:metrics:measure_magnitude}, Eq.~\eqref{eq:mag} \\
 
All channels
& \texttt{\{"anomalies": [\{"channel": int, "start": int, "end": int\}, \dots]\}} $\rightarrow$ event set $P$, greedily matched to the gold set $G$ (highest IoU first, same channel, $\mathrm{IoU} > 0.3$, each event used once)
& $2\,\mathrm{TP}/(2\,\mathrm{TP} + \mathrm{FP} + \mathrm{FN})$ over the matched events; an empty list scores $1$ on an anomaly-free row and $0$ otherwise
& Section \ref{app:anomalyxl:metrics:all_channels}, Eq.~\eqref{eq:setf1} \\
 
Lead--lag
& \texttt{\{"direction": str, "lag\_samples": int\}} $\rightarrow$ relation $\hat d \in \{\text{lead}, \text{lag}, \text{independent}\}$ and offset $\hat\ell$, reduced to $\delta = |\hat\ell - \ell|/L$
& $\mathbf{1}[\hat d = d]\cdot\max(0,\, 1 - \delta/\tau_\ell)$ with $\tau_\ell = 0.05$; the independent case has no gold offset and reduces to three-way classification
& Section \ref{app:anomalyxl:metrics:lead_lag}, Eq.~\eqref{eq:ll} \\
 
\midrule
 
\multicolumn{4}{@{}l}{\textit{\textsc{AnomalyXL-MCQ}} --- the raw final message, a \texttt{\{"answer": "\dots"\}} object, or a trailing \texttt{answer:} } \\
\addlinespace[1pt]
 
Presence
& Yes / No $\rightarrow$ selected option $\hat a$, by exact string match against the gold option text
& $\mathbf{1}[\hat a = a] \in \{0,1\}$
& Section \ref{app:anomalyxl:metrics:coarse_rows} \\
 
Start time
& Before / Early / Medium / Late, plus No Anomaly $\rightarrow$ $\hat a$
& $\mathbf{1}[\hat a = a] \in \{0,1\}$
& Section \ref{app:anomalyxl:metrics:coarse_rows} \\
 
Categorization
& Level Shift / Transient Spike / Change in Seasonality / Change in Variance / Change in Trend / No Anomaly. Four of the five kind labels (one dropped at random), plus No Anomaly $\rightarrow$ $\hat a$
& $\mathbf{1}[\hat a = a] \in \{0,1\}$
& Section \ref{app:anomalyxl:metrics:coarse_rows} \\
 
Magnitude band
& 1--3 / 3--10 / 10--30 / 30--100\,$\sigma$, plus No Anomaly $\rightarrow$ $\hat a$
& $\mathbf{1}[\hat a = a] \in \{0,1\}$
& Section \ref{app:anomalyxl:metrics:coarse_rows} \\
 
Identification
& Options for anomalous channels are two singletons, one pair, a triplet, and None, over three named channels $\rightarrow$ $\hat a$
& $\mathbf{1}[\hat a = a] \in \{0,1\}$
& Section \ref{app:anomalyxl:metrics:coarse_rows} \\
 
Indicator
& Leading / Lagging / Correlated / Uncorrelated / No anomaly $\rightarrow$ $\hat a$
& $\mathbf{1}[\hat a = a] \in \{0,1\}$
& Section \ref{app:anomalyxl:metrics:coarse_rows} \\
 
\bottomrule
\end{tabularx}
\caption{\textbf{AnomalyXL metrics across both benchmark formulations.} For each task, the output the model
is required to produce, the quantity recovered from it by the parser, and the resulting score.
Windows are half-open index intervals and $\mathrm{IoU}$ is defined in
Section \ref{app:anomalyxl:metrics:temporal_overlap}, Eq.~\eqref{eq:iou}. Gold quantities are unhatted, predictions hatted;
$L$ is the recording length. The two regimes are posed over the \emph{same} underlying
recordings, so the only difference between them is the scoring function in the third column:
continuous and evidence-grounded above, a $\{0,1\}$ choice indicator below. Per-category
numbers are unweighted row means (Section \ref{app:anomalyxl:metrics:aggregation}).}
\label{tab:anomalyxl_metrics}
\end{table*}

Each precise row is scored by a deterministic function of the model's final message
that returns a value in $[0,1]$. The same value is used as the evaluation score and,
during RL, as the reward.

\subsection{Temporal overlap.}
\label{app:anomalyxl:metrics:temporal_overlap}

Windows are defined as half-open index intervals $w = [s,e)$. For a predicted window
$\hat w = [\hat s, \hat e)$ and a gold window $w = [s,e)$,
\begin{equation}
\label{eq:iou}
\mathrm{IoU}(\hat w, w) \;=\; \frac{I}{(\hat e - \hat s) + (e - s) - I},
\end{equation}
where $I = \max\bigl(0,\, \min(\hat e, e) - \max(\hat s, s)\bigr)$ is the length of the
overlap. We set $\mathrm{IoU} = 0$ if the two intervals are disjoint or if either is
degenerate ($\hat s \ge \hat e$ or $s \ge e$). Every window-based task below uses this
definition.

\subsection{Localize ($r_{\text{loc}}$).}
\label{app:anomalyxl:metrics:localize}
Let $p, \hat p \in \{0,1\}$ be the gold and predicted presence flags.
\begin{equation}
\label{eq:loc}
r_{\text{loc}} =
\begin{cases}
1, & p = \hat p = 0,\\[2pt]
\mathrm{IoU}(\hat w, w), & p = \hat p = 1,\\[2pt]
0, & \text{otherwise.}
\end{cases}
\end{equation}
The second case requires a well-formed predicted window: a \texttt{present:\,true} answer
that omits \texttt{start} or \texttt{end} scores $0$. A presence mismatch in either
direction scores $0$.

\subsection{Classify with evidence ($r_{\text{cls}}$).}
\label{app:anomalyxl:metrics:classify_with_evidence}

The score is the product of a label term and a localization term,
\begin{equation}
\label{eq:cls}
r_{\text{cls}} = \mathbf{1}\!\left[\hat k = k\right] \cdot \mathrm{IoU}(\hat w, w),
\end{equation}
so a correct label earns credit only when it is paired with a window and the label acts as a scoring gate for the segment localization to score at all. The label
$k_{\varnothing} =$ ``No anomaly'' carries no window and 
set $r_{\text{cls}} = \mathbf{1}[\hat k = k = k_{\varnothing}]$. Labels are compared by exact
string equality against the five options given in the prompt.

\subsection{Measure magnitude ($r_{\text{mag}}$).}
\label{app:anomalyxl:metrics:measure_magnitude}

For gold peak deviation $m > 0$ and prediction $\hat m$, let $\rho = |\hat m - m| / m$.
The score is a linear ramp,
\begin{equation}
\label{eq:mag}
r_{\text{mag}} = \max\!\left(0,\; 1 - \frac{\rho}{\tau}\right), \qquad \tau = 0.5,
\end{equation}
which is $1$ at an exact match and $0$ at $50\%$ relative error. A non-numeric or absent
\texttt{magnitude\_sigma} sets $\rho = 1$ and hence $r_{\text{mag}} = 0$. Because $\rho$ is a
relative error and gold magnitudes are drawn log-uniformly in $[1,100]\,\sigma$, errors at the
low and high ends of that range are weighted equally.

\subsection{All channels ($r_{\text{ac}}$).}
\label{app:anomalyxl:metrics:all_channels}

This is the only task whose answer is a \emph{list} of events rather than a single window, so
the score compares two sets. Each event is a (channel, window) pair: the gold set
$G$ has $n_g$ of them and the predicted set $P$ has $n_p$.
 
We first decide which predictions found which gold events. A prediction counts as finding a
gold event if it names the same channel and its window has $\mathrm{IoU} > \theta$ with the
gold window, with $\theta = 0.3$. Pairs are formed greedily, highest IoU first, and each gold
event and each prediction is used at most once. Every pair formed this way is a true positive.
Predictions left unpaired are false positives, and gold events left unpaired are false
negatives. The score is the $F_1$ of those three counts,
\begin{equation}
\label{eq:setf1}
r_{\text{ac}} \;=\; \frac{2\,\mathrm{TP}}{2\,\mathrm{TP} + \mathrm{FP} + \mathrm{FN}},
\end{equation}
so a model is penalised both for missing events and for inventing them. On a row that
contains no anomalies at all, $r_{\text{ac}} = 1$ if the model returns an empty list and $0$
otherwise.

\subsection{Lead--lag ($r_{\text{ll}}$).}
\label{app:anomalyxl:metrics:lead_lag}

Let $d, \hat d \in \{\text{lead}, \text{lag}, \text{independent}\}$ and let $\ell, \hat\ell$ be
the gold and predicted offsets in samples. The offset earns credit only if the relation is
correct:
\begin{equation}
\label{eq:ll}
r_{\text{ll}} = \mathbf{1}\!\left[\hat d = d\right] \cdot
\begin{cases}
1, & d = \text{independent},\\[4pt]
\max\!\left(0,\, 1 - \dfrac{\delta}{\tau_\ell}\right), & \text{otherwise,}
\end{cases}
\end{equation}
where $\delta = |\hat\ell - \ell| / L$ and $\tau_\ell = 0.05$. The independent case has no
ground-truth offset, so it reduces to three-way classification. For the directional cases the
tolerance is a fraction of $L$, which keeps the task equally hard at every context length given
that gold offsets are drawn uniformly in $[1, L/4]$.

\subsection{MCQ tasks.}
\label{app:anomalyxl:metrics:coarse_rows}
Coarse (MCQ) rows are scored by exact string match of the extracted answer against the gold
option text, giving $r \in \{0,1\}$. Extraction accepts the raw final message, a JSON object
of the form \texttt{\{"answer":\,"\dots"\}}, or a trailing \texttt{answer:} line, in each case
after stripping any reasoning block. The paired design means the same recording is scored by
a continuous, evidence-grounded metric in the precise regime and by a $\{0,1\}$ choice
indicator in the coarse one, with nothing else differing between the two.

\subsection{Aggregation.}
\label{app:anomalyxl:metrics:aggregation}
Per-category numbers are the unweighted mean of the per-row score over the rows of that
category, and per-configuration numbers average those rows without reweighting by category
size. The all-channels column is therefore a macro average of Eq.~\eqref{eq:setf1} over rows,
not a micro-F1 pooled over global $\mathrm{TP}/\mathrm{FP}/\mathrm{FN}$ counts, and a
$K{=}1$ row carries the same weight as a $K{=}3$ row.


\section{Generalization Set Details}
\label{sec:appendix-generalization}

The generalization set samples from real clinical recordings---the Long-Term \ac{ECG} atrial-fibrillation
database~\citep{petrutiu2007ltaf} and the Sleep-PSG polysomnography
corpus~\citep{ghassemi2018you}---and are cast into the AnomalyXL-Localize \emph{classify-with-evidence} schema, so the same parser and scorers used throughout the benchmark apply unchanged. Each subset holds 250 windows: 200
carrying exactly one anomaly and 50 anomaly-free (Table~\ref{tab:generalization_stats}).

\begin{figure*}[t]
\centering
\includegraphics[width=\textwidth]{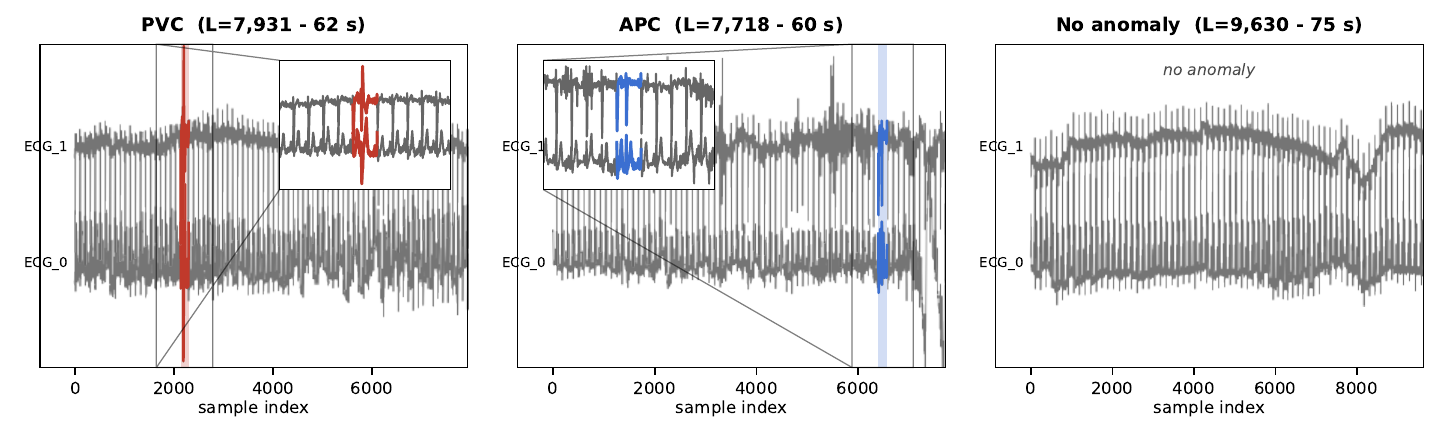}
\caption{\textbf{LTAF \ac{ECG} examples} (both channels, \texttt{ECG\_0}/\texttt{ECG\_1}, stacked): a PVC
beat, an APC beat, and a no-anomaly window, sampled from the released \texttt{ltaf-anomaly-v2} dataset.
Gold windows and exact values are given in the text above.}
\label{fig:ltaf-examples}
\end{figure*}

\begin{table}[h]
\centering
\small
\setlength{\tabcolsep}{3pt}
\begin{tabular}{@{}lll@{}}
\toprule
 & \textbf{LTAF (\acs{ECG})} & \textbf{Sleep-PSG} \\
\midrule
Rate, channels   & 128\,Hz, 2 & 200\,Hz, 4 \\
Windows          & 250 (200 / 50) & 250 (200 / 50) \\
Anomaly kinds    & PVC, APC & Central / Obstr.\ apnea, \\
                 &          & hypopnea, RERA \\
Anomaly          & one premature beat & one arousal bout \\
Gold window      & RR bracket & native bout range \\
Length (s)       & 60--1017 (med.\ 332) & 60--640 (med.\ 313) \\
Length (samples) & 7.7k--130k & 12k--128k \\
Gold span (med.) & $\approx$1.7\,s & $\approx$17.7\,s \\
Source records   & 9 LTAFDB & 150 PSG participants \\
\bottomrule
\end{tabular}
\caption{\textbf{Generalization set composition.} Window counts are given as
(anomalous / anomaly-free). Both subsets are balanced across anomaly kinds; channels are
z-normalized per channel.}
\label{tab:generalization_stats}
\end{table}
\paragraph{Worked examples: LTAF \acs{ECG}.}
Fig.~\ref{fig:ltaf-examples} plots sampled rows from the released LTAF subset at full resolution (thin
gray, both channels stacked) with the gold window highlighted and a zoomed inset. The PVC example
($L=7{,}931$, 62\,s) has gold \texttt{\{"kind": "Premature Ventricular Contraction (PVC)", "start": 2127,
"end": 2290\}}; the APC example ($L=7{,}718$, 60\,s) has gold \texttt{\{"kind": "Premature Atrial
Contraction (APC)", "start": 6396, "end": 6568\}}; the no-anomaly example ($L=9{,}630$, 75\,s) has gold
\texttt{\{"kind": "No anomaly"\}}. As in AnomalyXL itself, the gold span is a small fraction of the
window, so the full-resolution view is dominated by normal sinus rhythm and the ectopic beat only becomes
legible once the inset zooms in.

\begin{figure*}[t]
\centering
\includegraphics[width=\textwidth]{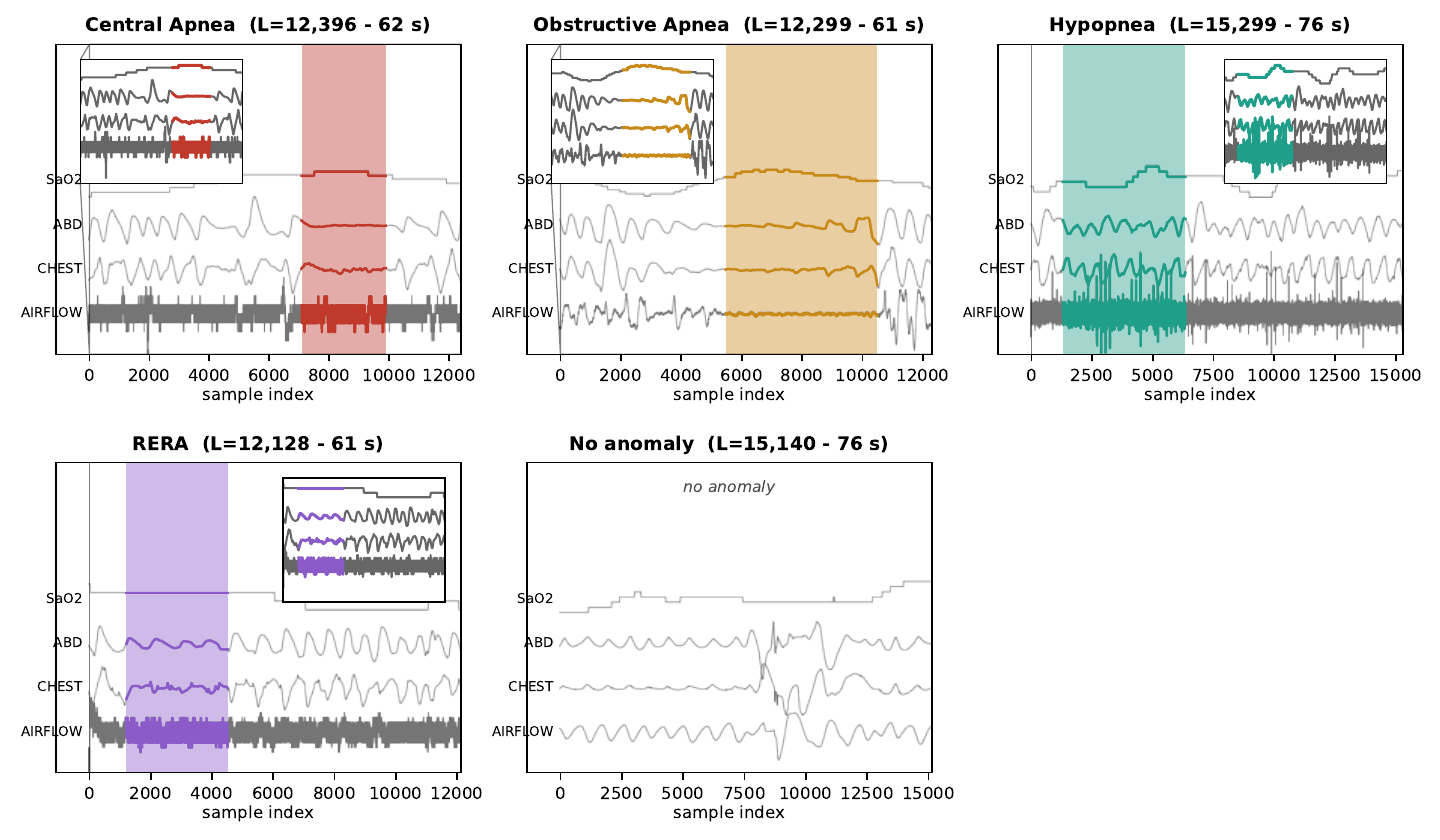}
\caption{\textbf{Sleep-PSG examples} (four channels, \texttt{AIRFLOW}/\texttt{CHEST}/\texttt{ABD}/\texttt{SaO2},
stacked): one window per respiratory-arousal kind plus a no-anomaly window, sampled from the released
\texttt{sleep-anomaly-v2} dataset. Gold windows and exact values are given in the text above.}
\label{fig:sleep-examples}
\end{figure*}

\paragraph{Worked examples: Sleep-PSG.}
Fig.~\ref{fig:sleep-examples} plots one sampled row per respiratory-arousal kind from the released
Sleep-PSG subset, at full resolution (thin gray, all four channels stacked) with the gold window
highlighted and a zoomed inset, each with its gold \texttt{[start, end)} bout window: Central Apnea
($L=12{,}396$, gold \texttt{start}=7080, \texttt{end}=9880), Obstructive Apnea ($L=12{,}299$,
\texttt{start}=5470, \texttt{end}=10490), Hypopnea ($L=15{,}299$, \texttt{start}=1300,
\texttt{end}=6340), RERA ($L=12{,}128$, \texttt{start}=1209, \texttt{end}=4509), and a no-anomaly example
($L=15{,}140$).

\begin{figure}[H]
\centering
\includegraphics[width=\linewidth]{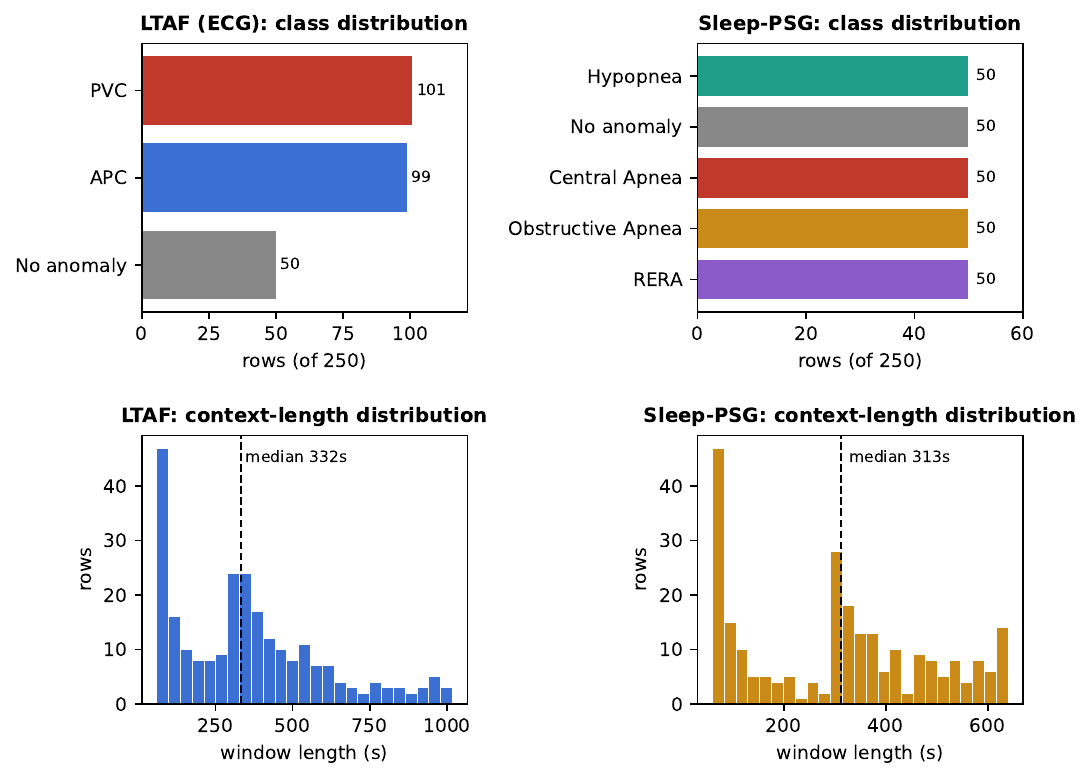}
\caption{\textbf{Generalization-set class and context-length distributions}, computed directly from the
released \texttt{ltaf-anomaly-v2} and \texttt{sleep-anomaly-v2} parquet files. Dashed line marks the
median window length.}
\label{fig:generalization-distributions}
\end{figure}

\paragraph{Window construction.}
Each positive window is drawn inside a corridor clipped at the nearest neighbouring event, which
guarantees that exactly one anomaly falls inside it. For LTAF this is a single premature beat---a
premature ventricular (PVC) or atrial (APC) contraction---inside a normal-sinus-rhythm bout containing no
other abnormal beat; negatives are clean sinus-rhythm spans. For Sleep-PSG it is a single respiratory
arousal---central apnea, obstructive apnea, hypopnea, or a respiratory-effort-related arousal---with no
other arousal in the window; negatives are arousal-free, non-wake spans. Within each corridor, the window
length itself is drawn uniformly across the full supported range: 60\,s up to $\approx$17\,min for LTAF,
60\,s up to $\approx$11\,min for Sleep-PSG, capped at $\approx$131k samples. Positive windows are drawn under a fixed per-kind quota (LTAF: PVC and APC each
$\approx$100 rows; Sleep-PSG: 50 rows per one of the four arousal kinds) via a round-robin scheduler that
alternates across kinds until each quota is met (Table~\ref{tab:generalization_stats}). Negative windows are drawn
independently from clean gaps of the same length range. The released windows come from two seeded passes
(42 for the base draw, 43 for a subsequent long-context extension) for reproducibility.
Fig.~\ref{fig:generalization-distributions} gives the resulting class balance and
context-length distribution: both length histograms are visibly bimodal --- a cluster of shorter windows
near the 60\,s floor and a broad tail of long-context windows reaching the $\approx$131k-sample cap --- so
short and long contexts are both well represented rather than concentrated near the median.

\paragraph{Unified detect-and-classify prompt.}
A single prompt template is used for both positive and negative windows, so detection and localization are
measured under one task rather than a separate present/absent question. The LTAF instantiation reads,
verbatim from the released dataset:

\begin{promptbox}[title={LTAF unified detect-and-classify prompt (verbatim, $L=7{,}931$)}]
Determine whether the time series contains an anomaly and, if so, identify it and its window. Respond with
exactly one JSON object on a single line, no other text: \{"kind": <label>, "start": <int>, "end": <int>\}.
\texttt{kind} must be one of: "Premature Ventricular Contraction (PVC)", "Premature Atrial Contraction
(APC)", or "No anomaly" if the series contains no anomaly. \texttt{start} is inclusive, \texttt{end} is
exclusive; if "No anomaly", omit them. Each anomaly kind is defined as follows: a Premature Ventricular
Contraction (PVC) is an early, wide and bizarre QRS complex arising from the ventricles\ldots{} (Here L =
7931.)
\end{promptbox}
The Sleep-PSG instantiation is identical in structure, substituting the four respiratory-arousal kind labels
and their clinical definitions (central apnea, obstructive apnea, hypopnea, RERA) for the \ac{ECG} beat
kinds. Scoring is \texttt{kind\_x\_iou}: kind-match times temporal IoU against the gold window, with a
correctly predicted \texttt{No anomaly} credited in full on negatives.

\end{document}